\documentclass[a4paper,fleqn]{cas-sc}

\usepackage[numbers]{natbib}

\usepackage{microtype}
\usepackage{graphicx}
\usepackage{algorithmic}
\usepackage{subfigure}
\usepackage{booktabs} 
\usepackage{dsfont}
\usepackage{placeins}
\usepackage{array,tabularx}
\usepackage{newunicodechar}
\usepackage{svg}
\usepackage{caption}   

\usepackage{algorithm}

\usepackage{hyperref}
\usepackage{xcolor}
\definecolor{orcidgreen}{RGB}{166,206,57}

\usepackage{enumitem}
\usepackage{lastpage}

\usepackage{amsmath}
\usepackage{amssymb}
\usepackage{mathtools}
\usepackage{amsthm}
\usepackage{pifont}
\newcommand{\cmark}{\ding{51}} 
\usepackage[capitalize,noabbrev]{cleveref}

\theoremstyle{plain}

\theoremstyle{definition}

\theoremstyle{remark}

\newtheorem*{mytheorem}{AI Harmonics}
\newenvironment{myproof}[1][Derivation of the metric]{%
  \par\noindent\textit{#1.}\quad\ignorespaces
}{\hfill$\qedsymbol$\par}
\newtheorem*{mycorollary}{Corollary}

\DeclareUnicodeCharacter{03C1}{\(\rho\)}

\usepackage[textsize=tiny]{todonotes}

\usepackage{threeparttable}

\begin{document}
\let\WriteBookmarks\relax
\def\floatpagepagefraction{1}
\def\textpagefraction{.001}
\shorttitle{AI Harmonics}
\shortauthors{S. Vei et~al.}

\title [mode = title]{AI Harmonics: a human-centric and harms severity-adaptive AI risk assessment framework}                   

\author[1]{Sofia Vei}[orcid=0009-0002-9906-7688,type=editor]
\cormark[1]
\ead{sofiavei@csd.auth.gr}

\credit{Conceptualization of this study, Methodology, Software}

\affiliation[1]{organization={School of Informatics, Aristotle University},
                postcode={54124}, 
                city={Thessaloniki},
                country={Greece}}

\author[2]{Paolo Giudici}[orcid=0000-0002-4198-0127]
\ead{paolo.giudici@unipv.it}

\credit{Conceptualization of this study, Methodology, Supervision}

\author[1]{Pavlos Sermpezis}[orcid=0000-0003-2129-977X]
\ead{sermpezis@csd.auth.gr}

\credit{Data curation, Writing - Original draft preparation}

\affiliation[2]{organization={Department of Economics and Management, University of Pavia},
                postcode={27100}, 
                city={Pavia},
                country={Italy}}

\author[1]{Athena Vakali}[orcid=0000-0002-0666-6984]
\ead{avakali@csd.auth.gr}

\author[3]{Adelaide Emma Bernardelli}[orcid=0009-0005-9271-9614]
\ead{adelaide.bernardelli@iusspavia.it}

\affiliation[3]{organization={University School for Advanced Studies - IUSS Pavia},
                postcode={27100}, 
                city={Pavia},
                country={Italy}}

\credit{Literature review, Database search}

\cortext[1]{%
  Corresponding author.  
  \textit{Address:} Ethnikis Antistaseos 16, 55133 Thessaloniki, Greece;  
  \textit{Tel.:} +30 2310991865
}
\begin{abstract}
The absolute dominance of Artificial Intelligence (AI) introduces unprecedented societal harms and risks. Existing AI risk assessment models focus on internal compliance, often neglecting diverse stakeholder perspectives and real-world consequences. We propose a paradigm shift to a human-centric, harm-severity adaptive approach grounded in empirical incident data. We present \textit{AI Harmonics}, which includes a novel AI harm assessment metric (\textit{AIH}) that leverages ordinal severity data to capture relative impact without requiring precise numerical estimates. \textit{AI Harmonics} combines a robust, generalized methodology with a data-driven, stakeholder-aware framework for exploring and prioritizing AI harms. Experiments on annotated incident data confirm that political and physical harms exhibit the highest concentration and thus warrant urgent mitigation: political harms erode public trust, while physical harms pose serious, even life-threatening risks, underscoring the real-world relevance of our approach. Finally, we demonstrate that \textit{AI Harmonics} consistently identifies uneven harm distributions, enabling policymakers and organizations to target their mitigation efforts effectively.
\end{abstract}


\maketitle

\section{Introduction}
Artificial intelligence (AI) is no longer just a tool; it is a force shaping life-altering decisions, yet it often does so at the cost of fairness, accountability, and human well-being. It is increasingly and rapidly embedded in critical decision-making processes and influences all critical domains, from healthcare and finance to employment and governance. Although AI presents immense opportunities, it brings about critical, unprecedented risks, and already different AI-related harmful incidents affect various stakeholders very differently. \textit{Harms in AI} refer to the actual negative impacts (such as physical, psychological, social, or economic loss) that occurred as a result of the deployment or misuse of AI systems. In contrast, \textit{risks in AI} refer to context-specific uncertainties, estimated by the possibility that a harm might occur in the future under certain conditions, reflecting the likelihood and potential severity of negative outcomes. Since organizations cannot address every potential risk at once, they must prioritize which AI harms need attention. \textit{Understanding both harms and risks} inter-relations and intensities is essential for responsible AI governance, since prompt risks detection will help anticipate and prevent future harms, while harms provide concrete evidence of where AI systems have already caused damage. 

All AI harms, whether caused, probable, implied, or speculative, are significant. However, in real-world cases, AI harms may vary significantly between various individuals, groups, applications, and fields. For example, AI harm may be straightforward, involving one AI system and a single harmed entity, but it can also be more complex and multifaceted, involving multiple harmed entities, AI systems, or types of harm. Furthermore, some AI harms may have too severe and irreversible consequences, hurting vulnerable groups based on their gender, age, disabilities, etc. \cite{avila2024early, jones2025don}. Thus, AI harms must be foreseen, managed, or mitigated early on and before any particular vulnerable individuals or groups are affected. Unfortunately, current risk prediction systems have been proven inefficient in several cases, such as in the Dutch welfare fraud case, which wrongfully determined that thousands of families were defrauding childcare benefits resulting in tangible harms (in the form of financial losses) but also in intangible harms (in the form of disparate treatment) for thousands of households \cite{Dutch-fraud}. Where AI harms already occurred, we may need to change the AI system and our AI risks mitigation processes. Unfortunately, 
not only the many different types of harm, but also their varying frequency and intensities make such an endeavor a rather challenging task. As these harm disparities become more evident, policymakers have begun reformulation of governance frameworks, with regulations such as the European AI Act \cite{smuha2024european, jariwala2024comparative}, which introduce varying risk-based management models to promote fairness, transparency, and accountability \cite{novelli2024taking}.

Until now, \textbf{AI risk assessment frameworks} have adopted an internal perspective, i.e., they focus on mitigating risks for AI providers and adopters through simple compliance measures \cite{raji2020closing, thomas2024case}. 
Such compliance measures explore only some limited registered risks, while mainstream internal audits and pre-defined checklists fail to recognize new and unforeseen harms. Such AI risk assessment frameworks largely overlook the perspective of those who experience harmful AI outcomes, i.e., users, communities, and all the adopters (external stakeholders) of AI solutions \cite{lancaster2024s}. 
Moreover, such frameworks are rigid and offer one-solution-fit-all tools which neglect both AI harms acuteness level and the domain-specific risks harshness.
Many frameworks apply the same set of generic fairness or privacy checks across all applications and fail to adjust to the differences in harm severity or domain nuances. For example, it is not efficient or credible to deliver identical recommendations to a medical diagnosis tool and a social media  service. Reliance on binary labels  (e.g., "Minor" vs. "Severe") can miss critical differences between a life‐threatening misdiagnosis and a minor ad bias. As AI-driven harms become more prevalent and intrusive \cite{jones2025don}, especially in critical domains such as health, finance, politics, there is an urgent need to replace and revise conventional corporate-based AI risk assessment with broader, stakeholder-oriented and harm-adaptive solutions.

\textbf{Assessing AI harms severity} in real‐world data often comes with inherent ambiguity and noise. Precise numerical scores - when they exist - are frequently arbitrary or inconsistently assigned, especially if AI harms annotations assigned in different and heterogeneous sources \cite{ braun2024beg}. For example, \cite{paeth2025lessons} report that curators in the AI Incident Database often assign different severity ratings to the same incident due to incomplete information and epistemic uncertainty, leading to inconsistent severity labels in all entries; similarly, \cite{davidson2017automated} show that crowd annotators disagree on whether nearly identical tweets constitute hate speech or simply offensive language, resulting in divergent numeric severity scores for the same content. Since many AI risks are unexpected, context-specific and embed subjective or hard-to-measure evidence, we lack sufficient data or quantitative measurements (eg. as probabilities of AI harms) \cite{raman2025intolerable, aven2018risk, flage2014concerns}. Given the scarcity of numerical evidence, we argue that it is important to introduce ordinal scales, which will enable AI risks to be ranked and prioritized.

To address these issues, we claim that a systematic approach must respond to the next critical Research Questions (RQs): \textbf{RQ1:} How can we reform AI risks assessment methods to embed human  perspectives about AI harms severity? \textbf{RQ2:} In the absence of precise numerical data, how can AI harms be meaningfully ranked and prioritized? \textbf{RQ3:} Can we design and deploy a framework that is \textit{robust} and \textit{adaptive} to diverse stakeholder inputs for assessing AI harm severity? We respond to the above RQs with the first  harm-centric and stakeholder-adaptive \textit{AI risk assessment framework} that embeds adopters views about AI harms captured over ordinal scales. Our main intuition for considering \textit{ordinal AI harms severity} as our default proposition is that it allows for a simple severity ordering and it doesn't depend on exact numerical severity distances and thus scales smoothly even when quantitative values are unavailable or unreliable. By relying solely on relative AI harms severity rankings, we avoid imposing unjustified harms assumptions and improve robustness under noisy or inconsistent annotations \cite{robinson2024likert}. The main ideas and contributions of the proposed approach are as follows.
\begin{itemize}[topsep=0pt]
    \item \textit{We embed human perspectives about AI harms}, by a \textbf{data-driven and consensus-based} approach which relies on human annotators’ judgments, delivered at incident‐tagging processes. Unlike existing AI risk management approaches that rely on internal risk assessments based on restrictive proprietary data (private user logs, proprietary performance metrics, internal testing, etc.), we evaluate risks from the perspective of external stakeholders to include greater input and insights from actual users or affected communities. Thus, \textit{in response to RQ1}, we integrate human perspectives by mapping text {\bf annotations} 
    collected from external stakeholders (e.g., affected end‐users, community representatives, and domain experts) to a selection of incidents reported in openly shared repositories. Our generalizable approach may  exploit any of the existing AI Harms related open data repositories, such as the AIAAIC Repository open database of incidents and controversies driven by AI, algorithms and automation~\cite{aiaaic}, which is chosen here due to its wide variety of incidents, extensive human annotation process and stable harm taxonomy on which it is based.

    \item \textit{We introduce a novel harm severity assessment approach}, which leverages a set of ordinal measurements ($\textit{AIH}$) to enable \textbf{AI harms ranking}. Our ordinal rankings $\textit{AIH}$ escape from the explicit (numerical) severity values which reflect the subjective (human perceived) AI harms intensity. In parallel, we introduce an AI harm type exploration by assessing the so called \textbf{AI harms severity concentration}, i.e., the estimation of inequalities among varying harms distributions. To do so,  we introduce a new metric $\textit{AIH}_{i}$ to assess each of the concentrations of harms $i$ and to allow a meaningful risk evaluation, in cases where the severity can only be classified ordinarily rather than measured numerically. Our approach covers cases where harms are mostly gathered into a few neighboring severity levels or distributed evenly across the entire scale. This allows for prompt intervention, and harm-focused adaptive mitigation strategies.
    These novel propositions respond to \textit{RQ2}, and the proposed metric is challenged against other metrics, such as the Criticality Index (CI), an ordinal risk assessment metric which has proven its capability to handle ordinal severity inputs, in cyber risk assessment settings \cite{giudici2021cyber}. 

    \item \textit{We design and introduce the AI Harmonics framework}, which allows for broad sensitivity analysis and demonstrates our proposed methodology and metrics suitability by extensive experimentation which showed that harm rankings remain stable under different severity orderings and annotation noise scenarios. AI Harmonics \textit{robustness} is supported by the fact that its harm‐category ordering remains consistent even when the size of the annotation set is changed, and its \textit{adaptiveness} is demonstrated by the framework’s ability to ingest revised annotations (e.g. annotators' severity judgments) and automatically recompute harm‐severity concentration metrics and rankings
    These checks directly address \textit{RQ3}, ensuring our approach remains reliable, efficient, and flexible as data and expert inputs evolve.
    
\end{itemize}

\noindent

In summary, AI Harmonics embeds stakeholder-informed annotations to directly capture harm impact on humans, while conventional frameworks often rely on internal, compliance-driven processes and omit empirical evidence from diverse stakeholders. Furthermore, AI Harmonics operates effectively on ordinal rankings, enabling meaningful prioritization without the need for precise numeric inputs. Lastly, our extensive sensitivity analysis demonstrates that AI Harmonics remains robust and adaptive distinguishing it from existing frameworks that offer static or one-size-fits-all solutions. Beyond ranking harms, our proposed framework enables a stakeholder-driven evaluation of AI risks, capturing nuanced harm scenarios that are often overlooked in broader risk management strategies. For instance, a governing body like the European Union (EU) could utilize this framework to prioritize harm mitigation efforts. Through expert annotation and consensus-building processes, opinions grounded in collected data can form the basis for such evaluations. This approach not only facilitates understanding of harm concentration across harm categories, but also provides actionable insights, such as identifying areas where controls are effective or mitigation is needed most urgently. 

The remainder of this paper is structured as follows. Section~\ref{sec:relatedwork} discusses prior research on AI risk management and fairness assessment, while Section~\ref{sec:fundamentals} introduces the core concepts to smoothly proceed with the main methodology of AI Harmonics. Section~\ref{sec:method} presents the proposed metric which drives the methodology and the proposed framework which are presented at Section~\ref{sec:framew}. Section~\ref{sec:results} provides an empirical analysis of harm severity distributions and Section~\ref{sec:conclusion} summarizes the findings and outlines future research directions.

\section{Related Work} \label{sec:relatedwork}
The widespread use of AI demands efficient \textbf{risk‐assessment regulatory models and frameworks} that must balance AI opportunities with AI risks \cite{le2022survey, groenewald2024smart, ashir2024impact}. Many such models align with recently adopted regulations - most prominently, the European (EU) AI Act - which set out high-level requirements for transparency, accountability, and fairness \cite{smuha2024european}. However, despite these comprehensive guidelines, most existing frameworks do not prescribe concrete data-driven methods to quantify or classify real-world harms, leaving practitioners to their own interpretations of 'high risk' vs. 'low risk' priorities. For example, Novelli et al. (2024) propose a scenario-based methodology to assess AI risk magnitudes by constructing real-world risk scenarios that complement the Act’s categories \cite{novelli2024ai}. Likewise, Smuha (2024) critically analyzes the EU AI Act’s potential to establish robust standards for AI systems \cite{smuha2024european}, and specialized, domain-specific work, such as Giudici (2023) in the financial sector, demonstrates how compliance to AI Act principles (Sustainability, Accuracy, Fairness, Explainability) can be formalized in risk management models. Recent advancements in AI governance emphasize that risk management must address both regulatory compliance and broader societal impacts \cite{ai2023artificial, qi2024ai, beckerstandardized, lim2025determinants}. For instance, the World Economic Forum’s AI Governance Alliance introduced a comprehensive framework centered on an end-to-end AI life cycle, robust risk safeguards, and early safety integration \cite{alliance2024briefing}. Similarly, Dehankar and Das \cite{dehankar2025ethics} stress the need to balance innovation with responsibility by proposing ethical guidelines that ensure technological advancement does not come at the expense of societal welfare. The AIR-Bench 2024 benchmark goes further by aligning AI safety evaluations with emerging regulations, offering a standardized framework for assessing AI risk across multiple domains \cite{zeng2024air}. Most AI risk management literature centers on the “internal” viewpoint, i.e., mitigating risks for AI providers and deployers by controlling in‐house processes and compliance checks \cite{hoffman2023outside, karnofsky2024developing, giudici2023lending, giudici2024artificial}. These frameworks focus on reducing liability or ensuring regulatory conformity but often treat harms as abstract categories rather than concrete impacts on real people. Consequently, they overlook the “external” viewpoint of affected communities - citizens, consumers, and vulnerable groups - who may have little control over or insight into AI deployments. In other words, existing approaches address AI risk management (internal controls) rather than AI risk awareness (public harms). This internal emphasis leaves unanswered questions about how to capture and prioritize the lived experiences of those harmed by AI systems, creating a critical gap in harm‐driven decision‐making.
Despite the growing sophistication of AI governance, systematic reviews consistently uncover persistent shortcomings, most notably around operationalization and meaningful stakeholder inclusion. For example, Novelli et al. and Xia et al.\cite{novelli2024taking, xia2023towards} conduct a systematic mapping of AI risk assessment frameworks to identify how well they align with responsible AI principles, integrate stakeholder considerations, and span different stages of the AI lifecycle. Their findings highlight significant gaps: \textit{few frameworks translate high-level risk categories into concrete, data-driven mitigation steps}, and even fewer embed affected communities’ perspectives into the assessment process. Additional surveys reinforce this conclusion, pointing out the lack of standardized procedures for converting qualitative risk terms into quantifiable priorities \cite{afzal2021review, golpayegani2022airo, kereopa2023safeguarding, ikudabo2024ai, novelli2024taking}. Although such earlier efforts push toward more comprehensive, society-focused risk models, they \textit{fail to offer a replicable method} to turn real-world incidents with diverse stakeholder voices into a prioritized list of concrete harms.

The Rise of \textbf{human-centric approaches in AI} has called for a shift toward stakeholder‐driven risk assessment, emphasizing the domains and communities most affected by AI harms \cite{chowdhury2024generative}. Raji and colleagues \cite{raji2020closing} propose an end‐to‐end algorithmic auditing framework that elevates transparency and inclusivity by bringing affected humans into the loop at each stage of the AI lifecycle. Participatory methodologies, where community representatives and domain experts help define harm criteria, seek to bridge the traditional divide between technical evaluation and lived experience. This “external viewpoint” prioritizes public experiences of harm over narrow, internal compliance checks. However, even these stakeholder‐centric approaches often \textit{lack a standardized mechanism} for aggregating and ranking multiple voices: in other words, they do not supply a clear, data‐driven way to measure how severe and concentrated different harms are once they arise in real incidents \cite{caton2024fairness}.
Building on more operationalized \textbf{AI risk metrics}, earlier work on various domain-specific and project-level cases has translated qualitative concerns into quantitative indicators. For example, a structured review mechanism adapts biosafety-style groupings - “AI-RG” (Artificial Intelligence Risk Groups) and “AS-RG” (Autonomous Systems Risk Groups) - to prioritize high-risk research areas (e.g., military or medical applications) via ordered risk tiers \cite{jordan2019designing}. In the cryptocurrency context, another model defines performance metrics (precision, recall, accuracy) to detect and rank AI-related financial threats \cite{elhoseny2023enhancing}. Yet another approach integrates regulatory constraints and business value into a “risk-aware actual value” score, extending the existing methodology so that organizations can balance innovation incentives against potential regulatory fines under frameworks like the AI Act \cite{ricciardi2023dilemma}. Although each of these methods offers useful tools to translate risk into numbers or scores, they tend to \textit{remain tied to narrow domains} or specific applications, and seldom incorporate a unified way to compare harms between disparate stakeholder groups. Shifting to an adaptive viewpoint demands alternative evaluation methods, since consensus on numeric risk measures is nearly impossible when harms range (eg. from mental distress to reputational damage). In analogous fields, such as cyber and operational risk, researchers have shown that ordinal scales can still yield reproducible prioritizations when quantitative data are scarce \cite{aldasoro2022drivers,liu2020unimodal}. Likewise, when the severity of a risk cannot be assigned a precise numerical value, measures such as the Gini concentration coefficient summarize the overall severity in a single index, even from purely ranked data \cite{giudici2021cyber,calzarossa2025robust}. Using ranked categories rather than exact scores, one captures relative severity among stakeholders without imposing arbitrary numerical assumptions. However, \textit{adopting ordinal scales alone is not enough} : first, is it really possible to generalize this methodology to any AI risk assessment? If so, how and which metrics are ideal to be used in this case? In addition, a robust analytical tool is required to convert these rank orders into an interpretable metric that reflects how the harms concentrate among the most affected. To address this gap, we therefore introduce an ordinal metric and benchmark it directly against the Criticality Index, showing how purely ranked severity data can be collapsed into a single, actionable score that highlights which harms demand the most urgent attention.

\paragraph{\textbf{Criticality Index as an Ordinal Benchmark}} The Criticality Index (CI) was originally devised for cyber risk modeling to quantify the severity of scenarios when only ordinal information is available \cite{facchinetti2018measure}. In essence, CI computes the average rank of each risk instance within a cumulative severity distribution, enabling prioritization without precise probability or loss figures. Some AI‐focused works have mentioned a “criticality index” concept under the EU AI Act, but only as a high‐level idea rather than a concrete implementation \cite{salim2024ensuring}. Thus, CI serves as a natural benchmark for any ordinal‐based harm metric because it produces a single score from ranked severity data. This extends to its use to validate $\textit{AIH}$ metric: since both CI and $\textit{AIH}$ (which uses a variation of Gini) reduce to linear transformations of one another when applied to the same ordinal rankings, a strong correlation between the two confirms that our method faithfully captures ordinal concentration of harms (see Appendix~\ref{lsec:vs} Table \ref{tab:pros_cons} for a detailed side-by-side breakdown of their respective strengths and weaknesses).

\paragraph{\textbf{Extending to Numerical Severity}} While CI is strictly ordinal, the classic Gini coefficient extends seamlessly to numerical severity values, making it a versatile tool whenever datasets supply explicit quantitative scores (e.g., the AI Incident Tracker’s numeric “ratings”). In contrast to CI’s rank‐average formula, Gini measures inequality by computing the area between the Lorenz curve and the line of equality. Importantly, when numeric severities are available, Gini provides a more granular view of harm concentration, whereas CI would collapse those numbers into ranks. Because $\textit{AIH}$ aligns with CI under pure ordinal inputs and converges to standard Gini under numerical inputs, it unifies both cases: whenever severity is inherently ordinal, it matches CI; whenever severity is numeric, it behaves like a classic Gini. This dual capability ensures that AI Harmonics can handle any dataset, ordinal or numerical, while preserving comparability and interpretability across datasets and stakeholder groups.

Only few partial solutions recognize all these gaps, and lack of a fully stakeholder-embedded, harm-focused approach. For instance, Ballot and colleagues \cite{ballot2025limitations} critique current regulatory efforts for being politically and market biased, noting that latent AI harms often escape formal measurement. Similarly, Schiff et al. \cite{schiff2024emergence} draw parallels between AI ethics audits and financial audits, but point out that even these ethics audits rarely incorporate external stakeholders or public reporting, limiting their practical usefulness. While these critiques underscore the urgent need for adaptable, continuously updated risk metrics, they stop short of supplying a concrete methodology that can systematically capture real-world incident data, represent a range of stakeholder viewpoints, and produce a ranked list of specific harms, precisely the gap our AI Harmonics framework is designed to fill.

Building on these insights and the identified gaps in existing frameworks, we introduce \textbf{AI Harmonics: a unified stakeholder-inclusive solution} that embeds real-world incident data, leverages ordinal severity rankings, and quantifies harm concentration to drive targeted AI risk mitigation. Firstly, it adopts the external viewpoint by using incident data annotated by humans, ensuring that harm scores reflect those who actually experience AI‐related damage rather than merely internal compliance metrics. Secondly, it leverages ordinal severity rankings, avoiding forced quantitative assignments, yet translates them into a tailored metric which measures how harm is concentrated among the most affected stakeholders. Thirdly, AI Harmonics validates its findings against an established ordinal metric, ensuring robustness and interpretability and it produces a ranked list of harm categories (and subcategories) based on real‐world severity distributions. Finally, this end‐to‐end methodology, from data ingestion to prioritization, is demonstrated under a proper framework to provide policymakers and practitioners with a reproducible, data‐driven solution for prioritizing AI harms that align with ethical principles and societal values.

\section{Fundamentals - Background} \label{sec:fundamentals}
This section introduces the core concepts and conceptual building blocks for our harm‐severity framework. We begin by formalizing key definitions which are essential for all the subsequent analysis. We then survey existing taxonomies or frameworks of AI harms and risks to position our work within the broader literature. Next, we outline the requirements that make a dataset suitable for our external, stakeholder‐focused methodology and we describe how affected \emph{stakeholder} groups and harm \emph{categories/subcategories} -the essential dimensions for the application of the framework- are represented in structured incident datasets. Together, these fundamentals establish the terminology, taxonomy, and data requirements that enable a consistent, dataset‐agnostic approach to quantifying and comparing the concentration of AI‐related harms.

Our hypothesis for treating the problem as an ordinal case is that using simple ranked categories by using stakeholder-defined severity levels, lets us capture a wider range of harm types more flexibly and accurately (e.g., "Minor"–"Moderate"–"Severe", represented by increasing numeric ranks). Accordingly, we introduce \textit{ordinal AI harms severity} as our default proposition since it requires only a total ordering, not exact distances among severity values, and thus scales smoothly even when quantitative values are unavailable or unreliable; by relying solely on relative rankings, we avoid imposing unjustified interval assumptions and improve robustness under noisy or inconsistent annotations \cite{robinson2024likert}. \begin{quote}
The advantage of ordinal severity scales (s) becomes clear when considering real-world annotation challenges. For instance, while annotators may reliably agree that \textit{financial loss} (severity=2) is worse than \textit{minor inconvenience} (severity=1) but not as severe as \textit{bodily harm} (severity=3), they would likely disagree on precise numerical intervals between these harms (e.g., whether the jump from 1→2 represents the same "distance" as 2→3). Ordinal scales accommodate this uncertainty by requiring only that $s_1 \leq s_2 \leq ... \leq s_M$ without assuming $s_2 - s_1 = s_3 - s_2$. This makes them particularly suitable for AI harm assessment, where severity judgments often come from multiple annotators with different perspectives \cite{davani2022dealing}, and where the "distance" between psychological, financial, and physical harms may not be meaningfully quantifiable.
\end{quote}

\subsection{Definitions}
In order to assess and prioritize AI-related harms systematically, we must begin by establishing a shared conceptual foundation. This subsection defines the core terms and constructs that underpin our framework, ensuring clarity and consistency throughout the methodology. Specifically, we formalize the notions of AI harm, harm concentration, AI risk, and severity, each of which plays a critical role in capturing how harm is experienced and distributed across different stakeholder groups. By anchoring our framework in these definitions, we enable a structured analysis of real-world incidents and a principled approach to comparing and ranking the severity of harms caused by AI systems.

\paragraph{\textbf{AI Harm.}}
An \textit{AI harm} is understood as physical, psychological, or otherwise consequential damage experienced by individuals, groups, organizations, society, or the environment as a result of the use or misuse of AI, algorithmic, or automated systems. These harms are characterized not only by their tangible or intangible nature but also by their scope, ranging from personal injury or reputational damage to large-scale societal or environmental degradation. A comprehensive taxonomy of such harms includes categories like autonomy loss, psychological trauma, reputational and financial damage, discrimination, and environmental impact, among others \cite{abercrombie2024collaborative}. According to OECD, the different dimensions of harm include: level of severity, scope, geographic scale, tangibility, quantifiability, etc. In this work we choose to deal with the level of severity as our main dimension of harms.

\paragraph{\textbf{AI Risk.}}
\textit{AI risk} is defined as a composite assessment of the probability that an AI system will produce an error, malfunction, or be exploited, multiplied by the expected impact of that event. It incorporates both the likelihood of adverse outcomes, such as model hallucinations, prompt injections, or biased predictions, and the severity of their consequences, which may span legal, operational, reputational, financial, and societal domains \cite{ai2023artificial}. This framework aligns with traditional risk assessment models used in other high-stakes domains and underpins regulatory approaches to AI safety and trustworthiness.

\paragraph{\textbf{(Harmful) AI Incident ($n$).}}
An \textit{AI incident} refers to an event, a set of circumstances, or a sequence of events in which the development, deployment, or malfunction of one or more AI systems results - either directly or indirectly - in one or more types of harm. Such harms may include: (a) physical injury or adverse effects on the health of individuals or groups; (b) disruption to the management or operation of critical infrastructure; (c) violations of human rights or breaches of laws protecting fundamental, labor, or intellectual property rights; and (d) damage to property, communities, or the environment. \cite{oecd2024aiincidents}.\\

\paragraph{\textbf{AI incidents dataset:}} To apply our framework, we need a dataset of AI incidents. For each incident we need to know to what AI system it corresponds (or the category of the AI system) and the severity of the incidents with respect to a harm --often related to a set stakeholders. For a clearer demonstration of the framework, we present the methodology and experimentation with respect with the AIAAIC dataset, which we describe in Section~\ref{sec:aiaaic}; however, the framework is generic and applicable to several other structures datasets, as we discuss in section~\ref{sec:suitability}.

We consider a dataset of $K$ harmful incidents, $n_{k}, k\in\{1, ...K\}$, each of which is associated with a stakeholder group ($h$) that is affected by the incident, and a harm category ($c$)  describing the nature of the harm.

\paragraph{\textbf{Stakeholders (h):}} Stakeholders are individuals, groups, or entities either directly or indirectly affected by the deployment and operation of AI systems. Stakeholders include diverse groups such as those who create content, develop and deploy technology, invest in AI systems, or simply interact with these technologies as part of their daily lives.\\
In the AIAAIC dataset that we use in this study, stakeholders are categorized according to a predefined taxonomy (see Appendix~\ref{lsec:appendix-data}), including: \textit{Artists/Content Creators, Business, General Public, Government/Public Sector, Investors, Subjects, Users, Vulnerable Groups, and Workers}.\\
We consider a set of $M$ stakeholders groups (such as \textit{users}, \textit{workers}, etc.), denoted as $h_{j}, j\in\{1, ...,M\}$. Stakeholders can be impacted/harmed either directly or indirectly by incidents relating to AI systems.

\paragraph{\textbf{Harm Category/Subcategory (c)}:} Harm categories represent broad areas of impact, while subcategories provide a finer-grained breakdown of specific harm types and they allow for a more detailed analysis of harm distribution and prioritization.\\
In the AIAAIC dataset, which follows a very thorough taxonomy \citep{abercrombie2024collaborative}, harm categories are structured into the following groups: \textit{Autonomy, Emotional \& Psychological, Financial \& Business, Human Rights \& Civil Liberties, Physical, Political \& Economic, Psychological, Reputational, Societal \& Cultural}. The subcategories specify granular classifications within these broader categories (see Appendix~\ref{lsec:appendix-data} for a complete list). This granularity allows for a more nuanced analysis of harm distribution.\\
We consider a set of $N$ high-level classifications of harm types (such as \textit{Psychological}, \textit{Financial}, or \textit{Reputational}) $c_{i}, i\in\{1,...,N\}$. Each category captures a unique dimension of harm.\\

\paragraph{\textbf{Frequency of stakeholders ($f$):}} For each harm category $c_{i}$, we calculate the conditional frequency of each stakeholder $h_{j}$ as:
\begin{equation*}
    f_{ij} = \frac{\sum_{k=1}^{K} \mathds{1}(n_{k}, c_{i}, h_{j})}{\sum_{k=1}^{K} \mathds{1}(n_{k}, c_{i})}
\end{equation*}
where $\mathds{1}(n_{k}, c_{i}, h_{j})$ is 1 iff the incident $n_{k}$ is associated with the harm category $c_{i}$ and the stakeholder $h_{j}$, and $\mathds{1}(n_{k}, c_{i})$,  is 1 iff the incident $n_{k}$ is associated with the harm category $c_{i}$ (irrespectively of the stakeholder).

\paragraph{\textbf{Severity (s):}}
\textit{Severity} refers to the magnitude or intensity of negative impacts resulting from the malfunction, misuse, or unintended consequences of an AI system. It is a key dimension in AI risk and harm assessment frameworks, used to evaluate how seriously an affected entity (individual, group, organization, society, or the environment) is harmed.\\
According to the MITRE Risk Discovery Protocol for AI Assurance, severity is formally defined as the magnitude of negative impacts on assets, operations, individuals, organizations, the Nation, or society more broadly \cite{ward2024risk}.\\
To each stakeholder group, we assign a severity value $s_{j}, j\in\{1,...,M\}$, which indicates the severity of the potential harms that an AI system can cause to the corresponding stakeholders. \textit{Without loss of generality, we assume that stakeholders are sorted based on their severity levels, i.e., $s_{j}<s_{j+1}, \forall j$.}

\begin{table}[ht]
\centering
\footnotesize
\caption{Notation Summary for AI Harmonics}
\label{tab:notation}
\begin{tabular}{p{2cm}p{10cm}}
\toprule
Symbol & Description \\
\midrule
$K$ & Total number of annotated incidents $n_k$, $k = 1,\dots,K$. \\
$N$ & Number of harm categories $c_i$, $i = 1,\dots,N$. \\
$M$ & Number of stakeholder groups $h_j$, $j = 1,\dots,M$. \\
$n_k$ & The $k$th incident in the dataset. \\
$c_i$ & The $i$th harm category. \\
$h_j$ & The $j$th stakeholder group. \\
$f_{ij}$ & Conditional frequency of stakeholder $h_j$ in category $c_i$: \[
    f_{ij} = \frac{\sum_{k=1}^{K} \mathds{1}(n_k, c_i, h_j)}
                    {\sum_{k=1}^{K} \mathds{1}(n_k, c_i)}.
  \] \\
$s_j$ & Ordinal severity ranking for stakeholder $h_j$, with $s_1 < s_2 < \dots < s_M$. \\
$\ell_i(x)$ & “Derivative” Lorenz curve for category $c_i$, induced by the severity ordering. \\
$\mathrm{AIH}_i$ & AI Harmonics concentration metric for category $c_i$: 
\[
  \mathrm{AIH}_i \;=\; \int_0^1 \ell_i(x)\,dx.
\] \\
$\mathrm{CI}_i$ & Criticality Index for category $c_i$, serving as an ordinal benchmark. \\
\bottomrule
\end{tabular}
\end{table}

Table~\ref{tab:notation} summarizes the main symbols used throughout AI Harmonics. We denote by $K$ the total number of annotated incidents $n_k$, each linked to one of $N$ harm categories $c_i$ and affecting one or more of $M$ stakeholder groups $h_j$. The relative frequency $f_{ij}$ captures how often stakeholder $h_j$ is harmed under category $c_i$. An ordinal severity ordering $s_j$ induces the “derivative” Lorenz curve $\ell_i(x)$, whose integral defines our AIH concentration metric $\mathrm{AIH}_i$. The Criticality Index $\mathrm{CI}_i$ serves as an ordinal benchmark. All of these definitions are introduced here for quick reference; they are developed in detail in Section~\ref{sec:method}.

\subsection{The AIAAIC dataset}\label{sec:aiaaic}
To demonstrate and evaluate our proposed framework, we utilize the AIAAIC dataset which documents a wide array of AI and algorithmic harm incidents. The dataset we analysed contains incidents recorded and annotated over a specific timeframe: from 2024/03/14 to 2024/04/11 - based on the structured taxonomy \footnote{In the analysis the alphabetical order is used, while we present all tables concerning categories ordered based on version v1.7 as per the AIAAIC harms taxonomy, adapted: e.g. "Environment" was not present in the dataset we were provided by the annotation tool, and thus we do not consider it as a stakeholder group. The definitions of the stakeholder categories present in the dataset we use are listed in Appendix~\ref{lsec:appendix-data} Table \ref{tab:stakeholders_definitions}.} of harms and affected stakeholders, as well as a carefully-designed specific methodology ~\cite{abercrombie2024collaborative}, a set of experts was invited to systematically review and annotate each incident. Annotators were responsible for examining incident descriptions and identifying relevant harm attributes, including the stakeholders affected, the harm category, and the subcategory of each incident.

Harm encompasses physical, psychological, or other types of damage inflicted on third parties due to the use or misuse of technology. The taxonomy also emphasizes that harm types are not mutually exclusive and may apply across multiple categories, reflecting the complex impacts of algorithmic systems.

In the dataset we were provided with, each incident was annotated independently by one or more experts, resulting in dataset of 816 annotations. Each annotation contains the following information: \textit{Datetime} of the annotation, \textit{Annotator ID}, \textit{Incident ID}, \textit{Stakeholders:} a group of stakeholders impacted by the incident, \textit{Harm Category/Subcategory}\footnote{In the AIAAIC harms taxonomy that we use, "Harm Categories" are referred to as "Harm Types" and "Harm Subcategories" are referred to as "Specific Harms". We use these terms to align with the terminology used in the annotated dataset of AIAAIC we were provided with.}: nature of the harm, \textit{Harm type:} which takes the values "actual" or "potential" and \textit{Notes:} optional field with comments of the annotator

The proposed framework needs only aggregate statistics about the stakeholders and harm categories. Specifically, we calculate the \textit{frequency} of stakeholders for each harm category, i.e., the number of annotations in the raw dataset that correspond to a given stakeholder group and category. In other words, we construct a dataset, where each row contains the following information: \textit{Harm Category (or Subcategory), Stakeholders, Frequency}.

For example:

\begin{center}
\begin{tabular}{ccc}
\toprule
\textbf{Harm Category} & \textbf{Stakeholder Group} & \textbf{Freq.} \\
\midrule
Autonomy & Artists/Content Creators & 15 \\
Autonomy & Workers & 5 \\
... & ... & ... \\
Societal \& Cultural & General Public & 29 \\
\bottomrule
\end{tabular}
\end{center}

In this structure, the \textit{frequency} captures how often a specific stakeholder group has been harmed within a given harm category. For example, the entry \textit{(Autonomy, Artists/Content Creators, 15)} means that there are 15 annotated incidents in the dataset where artists or content creators were harmed in ways classified under the \textit{Autonomy} harm category - such as loss of control over their personal data or unauthorized use of their work.
This table illustrates the minimal dataset structure required for applying our framework.

\begin{figure*}
\centering
\includegraphics[width=0.9\textwidth]{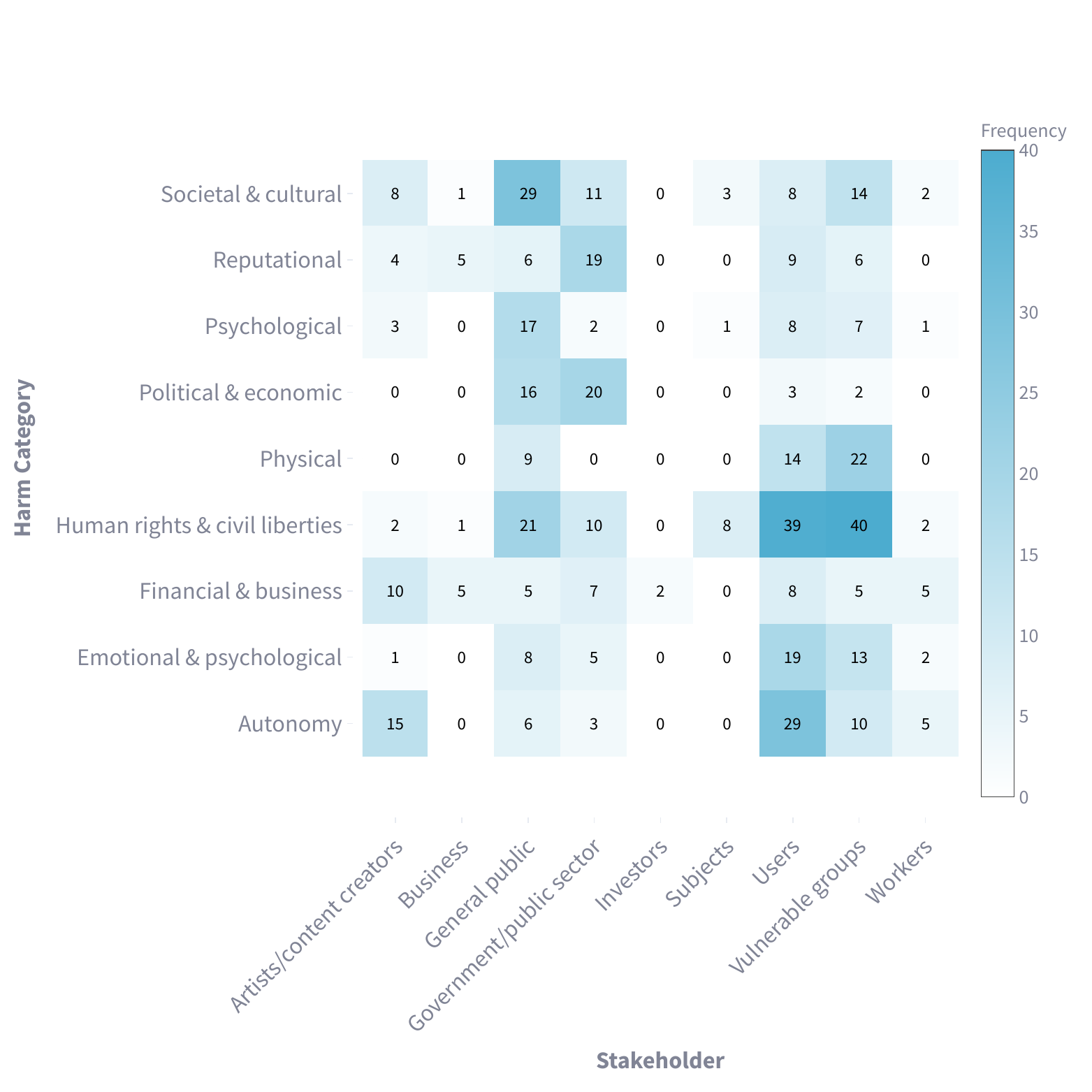}
\caption[Incidents by Harm Category \& Stakeholder]{\textbf{Incidents by Harm Category and Stakeholder Group}  
    Each cell shows the number of annotated incidents (\texttt{freq}) for a given harm category (rows)  
    and stakeholder group (columns). The color intensity (legend on the right) is also scaled to \emph{Frequency}.}
\label{fig:heatmap-frequency}
\end{figure*}

Figure~\ref{fig:heatmap-frequency} shows the frequency of annotated incidents for each harm category (rows) and stakeholder group (columns) for our experiments. The color intensity and overlaid numbers indicate how many times a given stakeholder was harmed within that category.  Notably, the \emph{Human rights \& civil liberties} category exhibits the highest counts - 40 incidents affecting \emph{vulnerable groups} and 39 affecting \emph{users} - underscoring a pronounced concentration of civil‐liberty harms on those two groups.  In the \emph{Autonomy} category, \emph{users} (29) and \emph{artists/content creators} (15) are most impacted, while in \emph{Physical} harms, \emph{vulnerable groups} (22) and \emph{users} (14) again bear the majority of incidents.  By contrast, stakeholders such as \emph{investors} and \emph{subjects} register zero or near‐zero incidents across almost all categories, indicating minimal exposure in our dataset.  Other categories, like \emph{Reputational} and \emph{Societal \& cultural}, show more moderate and somewhat even distributions (e.g.\ 19 reputational incidents for \emph{vulnerable groups}, 29 societal incidents for the \emph{general public}).  Overall, this heatmap makes clear that \textbf{users} and \textbf{vulnerable groups} disproportionately bear the most AI-related harms (in number), especially in human rights and civil liberties, autonomy, and physical well-being. For a complementary view of how the harm categories naturally group together based on their stakeholder‐impact profiles, see the hierarchical clustering dendrogram in Appendix~\ref{lsec:appendix-plotshsa} (Figure~\ref{fig:category_dendrogram}).

\section{$\textit{AIH}$: A new metric for AI harms assessment } \label{sec:method}
In this section, we construct and propose a metric to assess the harms of AI systems, which we denote as $\textit{AIH}$. 
We formally introduce the metric and present the process of its calculation that is inspired from the widely used Gini measure \cite{raffinetti2023rank}. $\textit{AIH}$ metric directly addresses the gap in adopting ordinal scales in a generalizable approach which could then be adapted to any AI risk assessment. In doing so, it enables a data-driven, stakeholder-inclusive prioritization of AI harms that bridges the divide between high-level fairness principles and concrete, incidence-based harm rankings.

We remind the reader that our goal is to classify categories of AI systems by the severity of their harms, to identify priorities of action in the oversight of AI systems. To this end, the $\textit{AIH}$ metric can be calculated for each category $c_{i}$ and then used to sort categories according to their risk of harms. The $\textit{AIH}$ metric takes values in the range $[0,1]$, with higher values indicating higher risks for a category of harm.
 
\begin{mytheorem}
    The $\textit{AIH}$ metric for a category $c_{i}$ with stakeholders $h_{j}, j\in{1,...,M}$ sorted by severity $s_{j}<s_{j+1}\forall j$ and having frequency $f_{ij}$, is defined as the integral
    \begin{equation}
        AIH_{i} =  \int_0^1 \ell_{i}(x) dx,
    \end{equation}
    where, for each category $c_i$,  $\ell_{i}$ is the curve defined by the points 
    \begin{equation*}
        \ell_{i} \rightarrow \left\{\sum_{j=0}^{k} f_{ij}, \frac{k}{M}\right\}, \forall k\in \{0,..., M\}
    \end{equation*}
where $f_{i0}=0$.
\end{mytheorem}
\begin{myproof}

The motivation of our method stems from the observation that harms caused by AI are not experienced uniformly. Some categories may distribute mild harm broadly, while others impose intense consequences on a few highly vulnerable groups. To capture this nuance, we need a metric that accounts for both severity and concentration. Classic inequality measures such as the Gini coefficient are well-suited for such distributional analysis. 

Hence, inspired by the Gini index, we extend it to the context of ordinal variables. $\textit{AIH}$ aligns with CI under pure ordinal inputs and converges to standard Gini under numerical inputs. This means that although here we highlight the ordinal cases, it addresses both cases: whenever severity is ordinal, it matches CI; whenever severity is numeric, the user can use Gini. We first compute the \textit{distribution ($f_{ij}$) of stakeholders $h_{j}$}. This distribution allows us to measure how harms in a specific category are allocated across different stakeholders.

The traditional Gini index takes values in the range $[0,1]$, with higher values indicating that most of the income is concentrated in a small set of individuals; and lower values indicating that the income is more evenly distributed. 

Here we need to replace income with severity, so that a high value of the Gini index means that, for a given category, most of the  severity is concentrated on a few high severity levels, and a low value means that the severity is more evenly distributed. Thus, categories with higher $\textit{AIH}$ will be prioritised as most harmful. 

However, this would hold only if severities were numerical (using Gini coefficient), that is, if we could assign numerical values to each stakeholder, in line with their order, using positive real numbers, $s_{i}\in\mathbb{R}$ such as  $s_{1}=1, s_{2}=2, ..., s_{M}=M$. Note that this assumes not only that stakeholders are ordered by the harm impact, but that such impacts grow linearly.  For example, if $s_{1}=1$ and $s_{2}=3$, then the severity for stakeholder $h_{2}$ is three times higher than for stakeholder $h_{1}$.

Normally, if severity levels had to be numerical values,  the Lorenz curve for a category $c_{i}$ would be defined by the following points:
\begin{equation*}
    L_{i} \rightarrow \left\{\sum_{j=0}^{k} f_{ij}, \frac{\sum_{j=1}^{k} f_{ij} \cdot s_{j}}{\sum_{j=1}^{M} f_{ij}\cdot s_{j}}\right\}, \forall k\in \{0,..., M\}
\end{equation*}
where we consider $f_{i0}=s_{0}=0$. The values of the points in the x-axis ($\textstyle\sum_{j=0}^{k} f_{ij}$) correspond to the cumulative frequency of the stakeholders, while the values of the points in the y-axis, correspond to the (normalized) cumulative share of severity of stakeholders. 

Then, the Gini index would be normally calculated as a function of the area below the Lorenz curve:
\begin{equation}
    G_{i} = 1 - 2\cdot \int_0^1 L_{i}(x) dx,
\end{equation}

In our case, severities cannot be assumed numeric. Hence, we need to modify the method for ordinal values.
One way to do so is to replace the Lorenz curve with its derivative or, in other words, the cumulative severity with the severity. Doing so, the resulting summary measure, that we call "pseudo-Gini" index, does not depend on the relationship between different severity levels which, as discussed before, may not be linear, but only on their order.

The "derivative" Lorenz curve $\ell_i$ for a category $c_{i}$ is defined by the $\{x,y\}$ points:
\begin{equation*}
    \ell_{i} \rightarrow \left\{\sum_{j=0}^{k} f_{ij}, \frac{k}{M}\right\}, \forall k\in \{0,..., M\}
\end{equation*}
where we consider $f_{i0}=s_{0}=0$ (i.e., the curve starts from the point $\{0,0\}$. The values of the points in the x-axis ($\textstyle\sum_{j=0}^{k} f_{ij}$) correspond to the cumulative frequency of the stakeholders, and the values of the points in the y-axis ($\textstyle\frac{k}{M}$) to the normalized severity value (since $s_{i}=i$ in this case), which we will refer to as "severity" from now on.

We can then calculate, in analogy with the previous case, a "pseudo-Gini" index,  the $\textit{AIH}$ metric, as the area below the derivative Lorenz curve:
\begin{equation}
    AIH_{i} =  \int_0^1 \ell_{i}(x) dx,
\end{equation}  
\end{myproof}

To further support our proposed metric for sorting categories of AI systems based on their harms severity, we compare it to a measure recently proposed in the cyber risk management literature: the \textit{Criticality Index} or $CI$.

The Criticality Index (CI) metric as used for ordinal cases for cyber‐risk modeling \cite{facchinetti2018measure}, computes the average rank of each scenario within its cumulative severity distribution, enabling prioritization without numeric probabilities or loss estimates. We use CI as a benchmark for our $\textit{AIH}$ metric because, when applied to the same ranked severity data, CI and $\textit{AIH}$ (pseudo-Gini) are mathematically equivalent up to a linear transformation. A strong empirical correlation between the two thus confirms that our $\textit{AIH}$ approach accurately captures ordinal harm concentration.

Specifically, when the impact of cyber attacks is expressed only in an ordinal scale, without numeric values, the Criticality Index (see e.g. \cite{facchinetti2019}), is defined as follows:
\begin{equation*}
    CI_{i} = \left\{ \frac{\sum_{j=1}^{M} F_{ij}-1}{M-1}\right\}, 
\end{equation*}
where $F_i^k=\textstyle\sum_{j=1}^{k} f_{ij}$,  for $k\in \{1,..., M\}$, indicate the values of the cumulative frequency distribution of the stakeholders. The CI is essentially the average of the values of the cumulative distribution.

It can be shown, by means of some simple mathematical steps, that our proposal and the CI as related: 

\begin{mycorollary}
For the $\textit{AIH}$ and the $CI$ metrics it holds that 
\begin{equation}
    AIH = CI \cdot \frac{(M-1)}{M} + \frac{1}{2\cdot M}
\end{equation}
\end{mycorollary}
This result further justifies the proposed index as a linear function of the CI, already known in the literature on ordinal cyber risk assessment, closely related to our context.

\begin{figure*}
\centering
\includegraphics[width=1\textwidth]{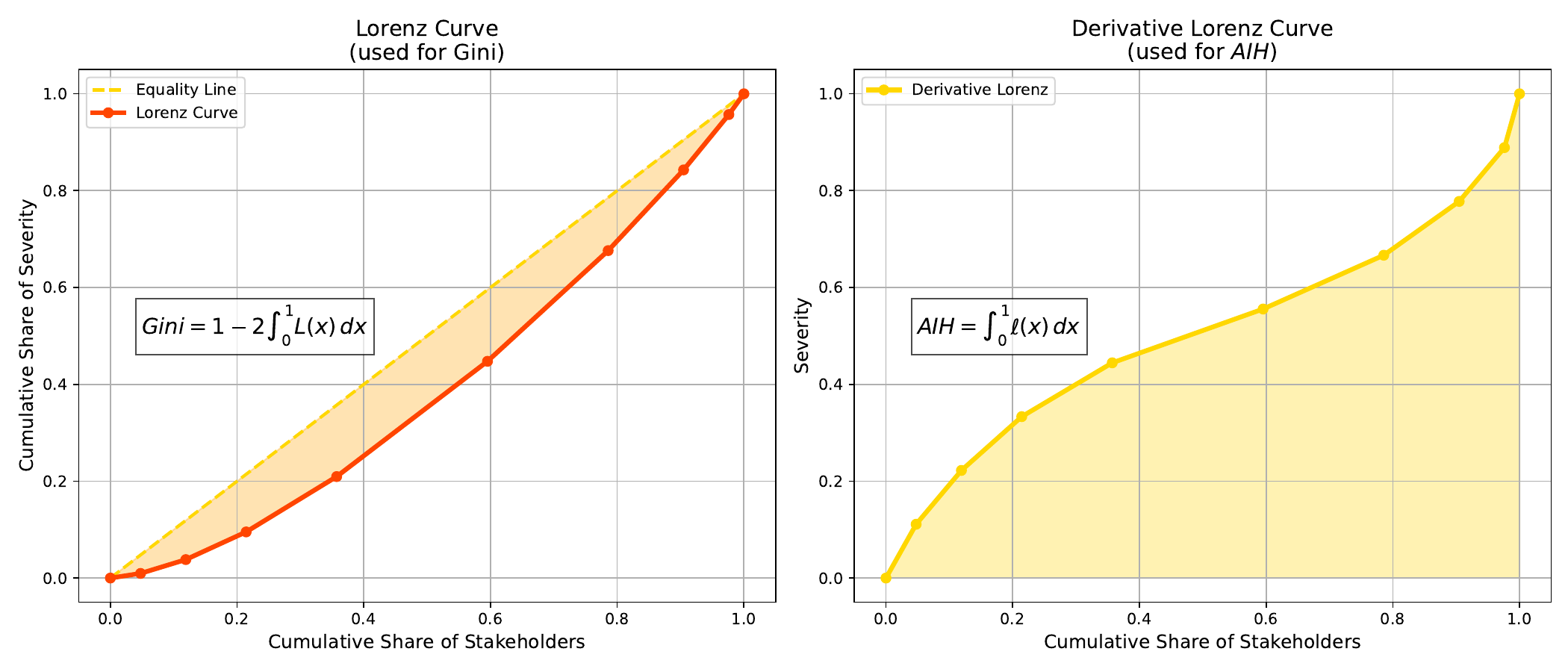}
\caption{\textbf{Left:} Classic Lorenz curve $L(x)$ that can be used when numerical severity is available.
    \textbf{Right:} “Derivative” Lorenz curve \(\ell(x)\) that can be used when only ordinal severity is available.}
\label{fig:lorenz_vs_pseudogini}
\end{figure*}

Figure~\ref{fig:lorenz_vs_pseudogini} contrasts the two constructions. On the left, the classic Lorenz curve shows how inequality is measured by the area between the 45° equality line and $L(x)$ (hence the factor $1-2\int_0^1L(x)\,dx$). On the right, the derivative‐Lorenz curve $\ell(x)$ plots severity ranks directly against cumulative population, and its entire shaded area \(\int_0^1\ell(x)\,dx\) defines $\textit{AIH}$. This side-by-side view makes clear that Gini quantifies the \emph{departure from equality} of a distribution of \emph{cumulative} severities, and $\textit{AIH}$ measures the \emph{aggregate severity} of an \emph{ordinal} ranking without ever assuming numeric intervals.

With both formulas in hand and visually compared, we can now move on to the overall AI Harmonics pipeline in Section~\ref{sec:framew}.

\section{The AI Harmonics Framework and Methodology} \label{sec:framew}
This section presents the core methodological framework proposed for assessing the severity of AI-related harms in a systematic, stakeholder-oriented, and data-driven manner. At its foundation, the framework introduces a pipeline for transforming incident-level data into actionable insights regarding harm concentration across various categories. Unlike conventional AI risk assessments that rely on predefined metrics or internal system parameters, our approach emphasizes the external perspective, incorporating information about who is harmed, how, and to what relative extent. Essentially, this Section combines and briefs the previous sections by introducing (i) the whole process -even before the annotation of a raw dataset, to the identification of the most harmful categories- and, (ii) how the whole methodology can be generalized to be applied in any dataset, as previously described.

The framework is designed to be both flexible and generalizable. It accommodates structured incident datasets with varying levels of granularity, supporting analyses that range from stakeholder-specific harm distributions to broader, category-level evaluations. Crucially, it does not require cardinal (numeric) severity assessments; instead, it operates on ordinal rankings - a minimal yet powerful input assumption. This design choice ensures applicability even when only comparative judgments about severity are available, which is often the case in real-world policy, ethics, or expert-driven contexts.

The overall goal of the framework is to enable the prioritization of harm categories based on how concentrated the most severe harms are. This concentration - or inequality in harm distribution - is captured through two complementary metrics: \textbf{\textit{AIH}}, and the \textbf{CI}, both of which reflect how harm is distributed across stakeholders or other relevant units. Both allow us to determine which types of harms should be considered most urgent or severe, based on their distributional patterns rather than raw frequency counts.

\begin{figure*}
\centering
\includegraphics[width=1\textwidth]{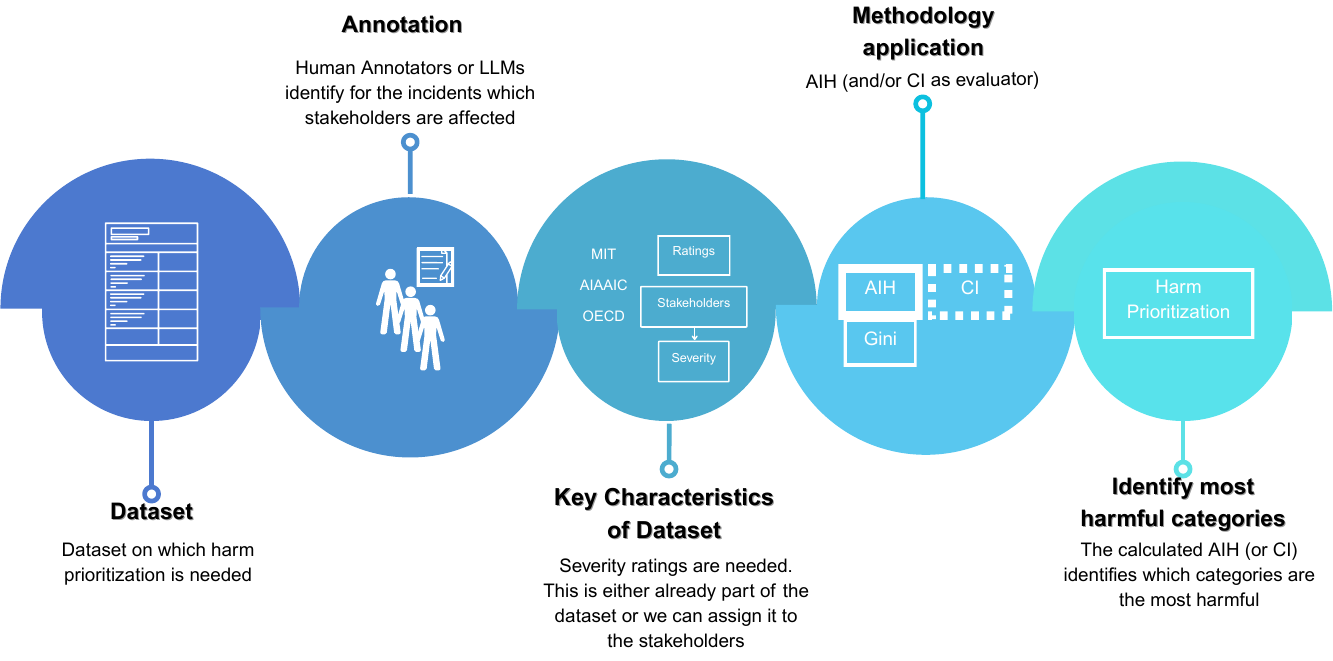}
\caption{Overview of the proposed stakeholder-oriented AI harm assessment framework, illustrating the main steps from dataset input to harm prioritization.}
\label{fig:framework}
\end{figure*}

Figure~\ref{fig:framework} provides a high-level visualization of the framework, which consists of five main stages: (1) dataset selection and suitability checks, (2) stakeholder annotation (if applicable), (3) severity ordering, (4) application of ordinal inequality measures, and (5) interpretation and prioritization of results, while Table~\ref{tab:framework_pipeline} briefs the following paragraphs which describe each stage in detail.

\paragraph{Dataset Selection and Suitability.}
The process begins by identifying a dataset of AI-related incidents for which harm prioritization is desired. The dataset must include structured records that capture at minimum a Harm Category for each incident. Ideally, datasets should also contain annotations on affected stakeholder groups and information about harm severity. When these are missing, human experts or automated methods (e.g., LLMs) may be used to enrich the dataset with such annotations.

\begin{table}[ht]
\centering
\footnotesize
\caption{Summary of the AI Harmonics Framework Pipeline}
\label{tab:framework_pipeline}
\begin{tabularx}{\textwidth}{p{3cm}X}
\toprule
\textbf{Stage} & \textbf{Description} \\
\midrule
Dataset Selection & Identify a structured incident dataset. Preferably includes annotations on harm category, stakeholder group, and severity (ordinal or numeric). \\
Stakeholder Annotation & If not already annotated, apply manual or automated annotation to assign stakeholder groups to each incident. Enables stakeholder-centric severity analysis. \\
Severity Ordering & Define ordinal severity levels across stakeholder groups (or categories if stakeholders are unavailable). This ordering can be expert-driven or participatory. \\
Metric Application & Apply the $\textit{AIH}$ metric (and/or the CI as a validator) to each harm category to compute concentration scores based on ordinal inputs. \\
Harm Prioritization & Rank harm categories based on computed concentration scores. Higher values indicate more severe concentration of harm and greater urgency for intervention. \\
\bottomrule
\end{tabularx}
\end{table}

\paragraph{Stakeholder Annotation.}
Next, domain experts or large language models are employed to determine which stakeholders were affected in each incident. These stakeholder annotations enrich the dataset by associating real-world harms with impacted societal groups, enabling a more grounded and socially relevant harm assessment. This step ensures the framework remains adaptable to various types of input data, including those lacking detailed stakeholder metadata.

\paragraph{Severity Rating Assignment.}
To assess the concentration of harm, each record - whether defined at the stakeholder level or directly at the incident or harm category level - must be situated along a severity scale reflecting the relative impact of harm. Our methodology requires only an ordinal assessment of severity: records must be ranked according to the relative severity of harm they represent, without requiring explicit numerical values. This ranking captures comparative vulnerability or criticality, depending on the level at which the data is structured. When stakeholder annotations are available, the ranking applies to stakeholder groups. When such information is absent, the ordering applies directly to the incidents or to harm categories. In our experiments, we illustrate this process by assigning severity scores to create a total ordering among stakeholders. However, these scores serve only as a means to establish the rank - our analysis is agnostic to the actual numeric values and relies solely on the ordering. This flexibility ensures the method remains broadly applicable: as long as an ordinal ranking - derived from expert judgment, context, or heuristics - is available at the appropriate level, the proposed methods can be applied effectively.

To make the flow from raw dataset to prioritized harms clear, we summarize the entire five-stage framework in Algorithm~\ref{alg:ai_harmonics}. It walks through data prep, stakeholder tagging, severity ordering, inequality
measurement and final ranking, showing exactly how to turn an annotated incident repository into a ranked harm list.

\begin{algorithm}[H]\small
\caption{AI Harmonics}\label{alg:ai_harmonics}
\begin{algorithmic}[1]
\REQUIRE Structured incident dataset with:
    \STATE \textbf{Harm Categories}: $C = \{c_1, ..., c_N\}$
    \STATE \textbf{Stakeholders}: $H = \{h_1, ..., h_M\}$ (optional)
    
\ENSURE Prioritized list of harm categories by concentration
    
\STATE \textbf{Step 1: Data Preparation}
    \IF{Stakeholder information available}
        \STATE Annotate incidents with $(c_i, h_j)$ pairs
    \ELSE
        \STATE Treat each incident as separate annotation
    \ENDIF
    
\STATE \textbf{Step 2: Severity Ordering}
    \STATE Establish ordinal ranking $s_j$ of stakeholders $h_j \in H$ (or incidents):

    \STATE $s_1 \leq s_2 \leq ... \leq s_M$ (where $\leq$ indicates "less severe than")
    
\STATE \textbf{Step 3: Distribution Calculation}
    \FOR{each harm category $c_i \in C$}
        \STATE Compute frequency distribution across stakeholders:
        \STATE $f_{ij} = \frac{\text{count}(c_i,h_j)}{\sum_{j=1}^{M} \text{count}(c_i,h_j)}$
    \ENDFOR
    
\STATE \textbf{Step 4: Concentration Measurement}
    \FOR{each harm category $c_i \in C$}
        \STATE Calculate $\textit{AIH}$:
        \STATE Construct derivative Lorenz curve $l_i(x)$
        \STATE $G'_i = \int_0^1 l_i(x)dx$
    \ENDFOR
    
\STATE \textbf{Step 5: Harm Prioritization}
    \STATE Rank categories by $G'_i$
    \STATE Higher values indicate more concentrated/severe harms
    
\STATE \textbf{Output}: Prioritized list $\{(c_i, G'_i)\}_{i=1}^N$
\end{algorithmic}
\end{algorithm}

\paragraph{Metric Application.}
Once we have the severity ordering - either across stakeholder groups or directly over incidents or harm categories - the framework applies one of two complementary metrics: the \textbf{\textit{AIH}} metric or the \textbf{CI}. Both approaches operate under the assumption of ordinal severity, requiring only a relative ranking of harm rather than explicit numeric values. The $\textit{AIH}$ metric derives from severity ranks to quantify the inequality - or concentration - of harm within each category. Similarly, the CI measures the average position of stakeholders (or records) in the cumulative distribution of harm severity. In both cases, a higher score indicates that harm is more concentrated among those most severely affected, and thus signals a more critical harm pattern. These metrics are suitable for any dataset in which a consistent ordinal ranking of severity can be established, regardless of whether stakeholder-level annotations are present.

\paragraph{Harm Prioritization and Interpretation.}
Finally, harm categories are ranked based on their calculated values using either the \textbf{\textit{AIH}} or the \textbf{CI}, both of which capture the concentration of harm severity. Higher index values indicate categories where harms are disproportionately concentrated, signaling the need for greater attention. These categories are flagged as priorities for regulatory action, policy response, or further investigation. By quantifying how harm is distributed across severity levels - whether or not stakeholder-level information is available - the framework supports a systematic, data-driven approach to identifying the most critical areas of AI-related risk.

\section{Experimentation} \label{sec:results}
This section presents a comprehensive analysis of the harm severity distribution across different stakeholder groups impacted by AI-related incidents (Section~\ref{subs: hsa}). After presenting the fundamentals, overview and preprocessing of our benchmark dataset in Section~\ref{sec:fundamentals}, and the main metric methodology in Section~\ref{sec:method}, by examining the \textit{AIH} metric, Lorenz curves, the CI and several plots, this analysis seeks to uncover the degree of inequality in harm distribution and identify which harm categories experience the highest levels of disparity (Section~\ref{ginicumprob}).

Additionally, we perform an extensive sensitivity analysis (Section~\ref{sec:sensitivity-analysis}), to explore the robustness of the proposed metric, under variations of data and perturbations that may appear in real settings.

\subsection{Harm Severity Analysis} \label{subs: hsa}
The harm severity analysis centers on examining harm distribution across the high-level \textit{Categories} of harm. (A detailed breakdown by both \textit{Categories} and more specific \textit{Subcategories} is provided in Appendix Table~\ref{tab:categories_subcategories}). The goal of the analysis is to evaluate how harm is distributed within each category, whether it is concentrated among those most severely affected or spread more evenly, in order to rank the categories accordingly.

To perform this analysis, we must first establish a severity ranking over the stakeholder groups. This ranking reflects the relative seriousness of harm when it affects different stakeholders. Our framework requires only an ordinal ordering - that is, a consistent sequence from least to most severely affected - and not specific numeric severity values. However, for the purposes of experimentation and illustration, we assign numeric values to stakeholders (as shown in Table~\ref{tab:stakeholders_severity}) that reflect such an ordering.

It is important to stress that the actual numbers assigned are inconsequential: only the order they represent matters. For instance, assigning a severity of 9 to the "General public" group versus 10 would not change the results of either the $\textit{AIH}$ or the CI metric. This is because both measures operate on ordinal information - they are sensitive to the ranking of stakeholders, not to the magnitude of differences between them. In fact, the same analytical outcomes would result from a purely symbolic ordering such as:\\
\textit{Artists/Content Creators} < \textit{Subjects} < \textit{Business} < \textit{Investors} < \textit{Workers} < \textit{Users} < \textit{Vulnerable Groups} < \textit{Government/Public Sector} < \textit{General Public}

The numeric values in Table~\ref{tab:stakeholders_severity} thus serve only as a means to encode this order and make the procedure more transparent and reproducible. In practice, users of the framework may specify this ordering directly or rely on domain expertise, contextual judgment, or participatory inputs to define it.

\begin{table}
\centering
\footnotesize
\caption{Stakeholder groups and their corresponding severity levels used to induce an ordinal ranking (for illustrative purposes only). The numeric values are arbitrary and serve solely to encode the order from least to most severe. The proposed methodology relies only on the ranking of these values, not their magnitude, and the ordering can be fully customized by the user.}
\begin{tabular}{|p{4cm}|p{2cm}|}
\hline
\textbf{Stakeholders} & \textbf{Severity} \\
\hline
Artists/content creators & 1 \\
\hline
Subjects & 2 \\
\hline
Business & 3 \\
\hline
Investors & 4 \\
\hline
Workers & 5 \\
\hline
Users & 6 \\
\hline
Vulnerable groups & 7 \\
\hline
Government/public sector & 8 \\
\hline
General public & 9 \\
\hline
\end{tabular}
\tiny
\label{tab:stakeholders_severity}
\end{table}

To account for possible differences in severity interpretations, we also include an extensive sensitivity analysis (Section~\ref{sec:sensitivity-analysis}) to test how robust our results are under alternative rankings and perturbations. This ensures that our prioritization of harm categories is not an artifact of any specific severity assignment, but rather reflects generalizable patterns of harm concentration.

\subsection{Summary measures} \label{ginicumprob}
We first apply $\textit{AIH}$ which allows to identify, for each category, which are the most frequent severities, and where they are located: at the beginning, middle, or end of the severity scale. 

In this approach, we calculate $\textit{AIH}$ based on severities, with severity thresholds representing the proportion of stakeholders affected at or below each severity level. This method allows to determine the intensity of harms across the spectrum of stakeholder groups, analyzing how many stakeholders experience harm up to specific severity levels.

\begin{figure*}
\centering
\includegraphics[width=0.8\textwidth]{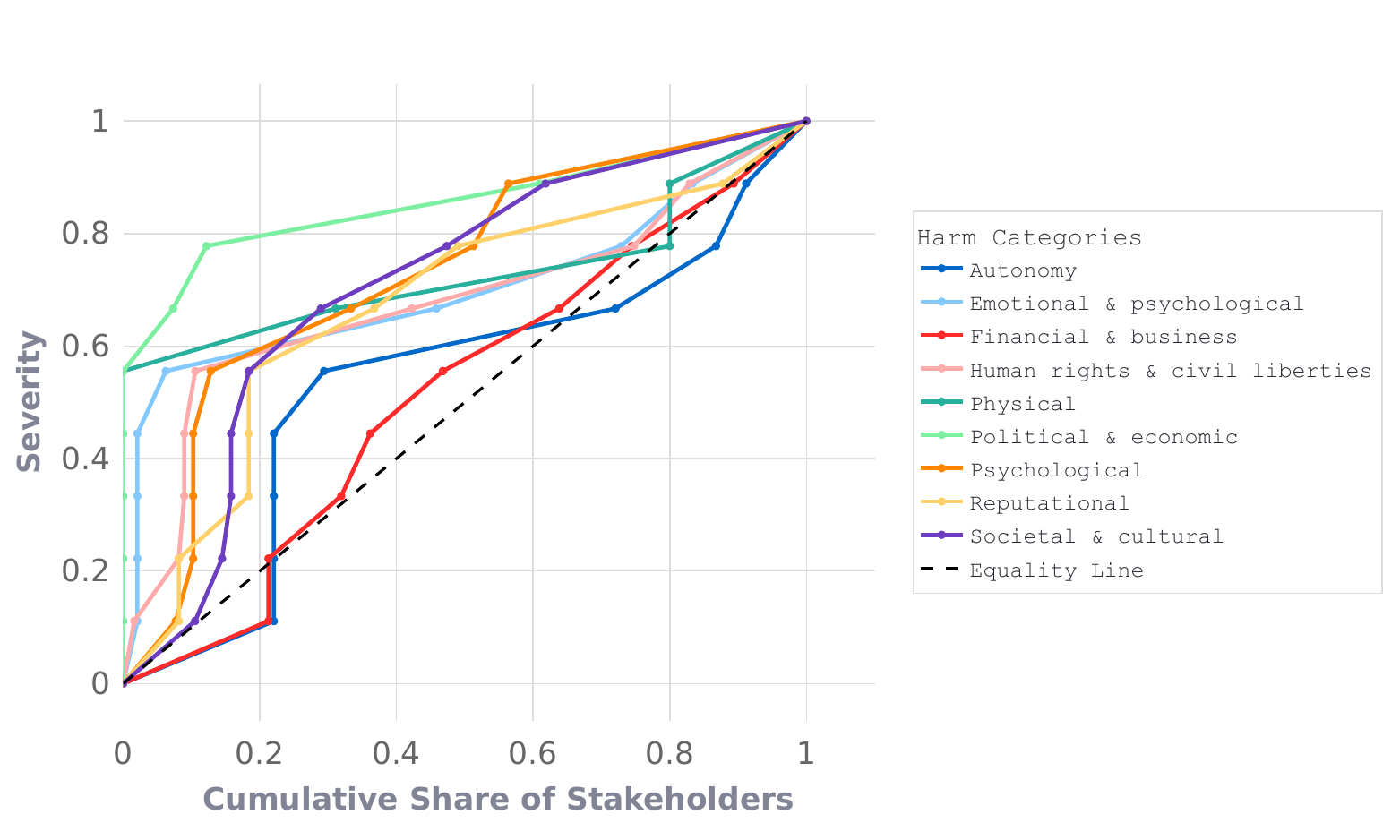}
\caption{Lorenz Curves for Harm Categories based on Severity}
\label{lorenzb}
\end{figure*}

Figure \ref{lorenzb} shows that the harm categories \textit{Financial \& Business} and \textit{Autonomy} have the lowest curves: in both cases, and especially for the former, the frequency of stakeholders is quite uniform across the nine considered classes.  Conversely,  the \textit{Political \& Economic} harm category appears to be highly concentrated on the highest classes.  The latter will lead to a high value of $\textit{AIH}$: more dangerous; and the former to a low value: less dangerous.

\begin{table}[ht]
\centering
\footnotesize
\caption{\textit{AIH} and CI values across harm categories}
\begin{tabular}{|l|c|c|}
\hline
\textbf{Harm Category} & \textbf{\textit{AIH}} & \textbf{CI} \\
\hline
Autonomy & 0.53 & 0.54 \\
Emotional \& psychological & 0.70 & 0.72 \\
Financial \& business & \textbf{0.51} & 0.51 \\
Human rights \& civil liberties & 0.67 & 0.70 \\
Physical & 0.73 & 0.76 \\
Political \& economic & \textbf{0.85} & 0.89 \\
Psychological & 0.73 & 0.75 \\
Reputational & 0.67 & 0.69 \\
Societal \& cultural & 0.70 & 0.73 \\
\hline
\end{tabular}
\label{tab:gini-criticality-index}
\end{table}

\begin{figure*}
\centering
\includegraphics[width=0.8\textwidth]{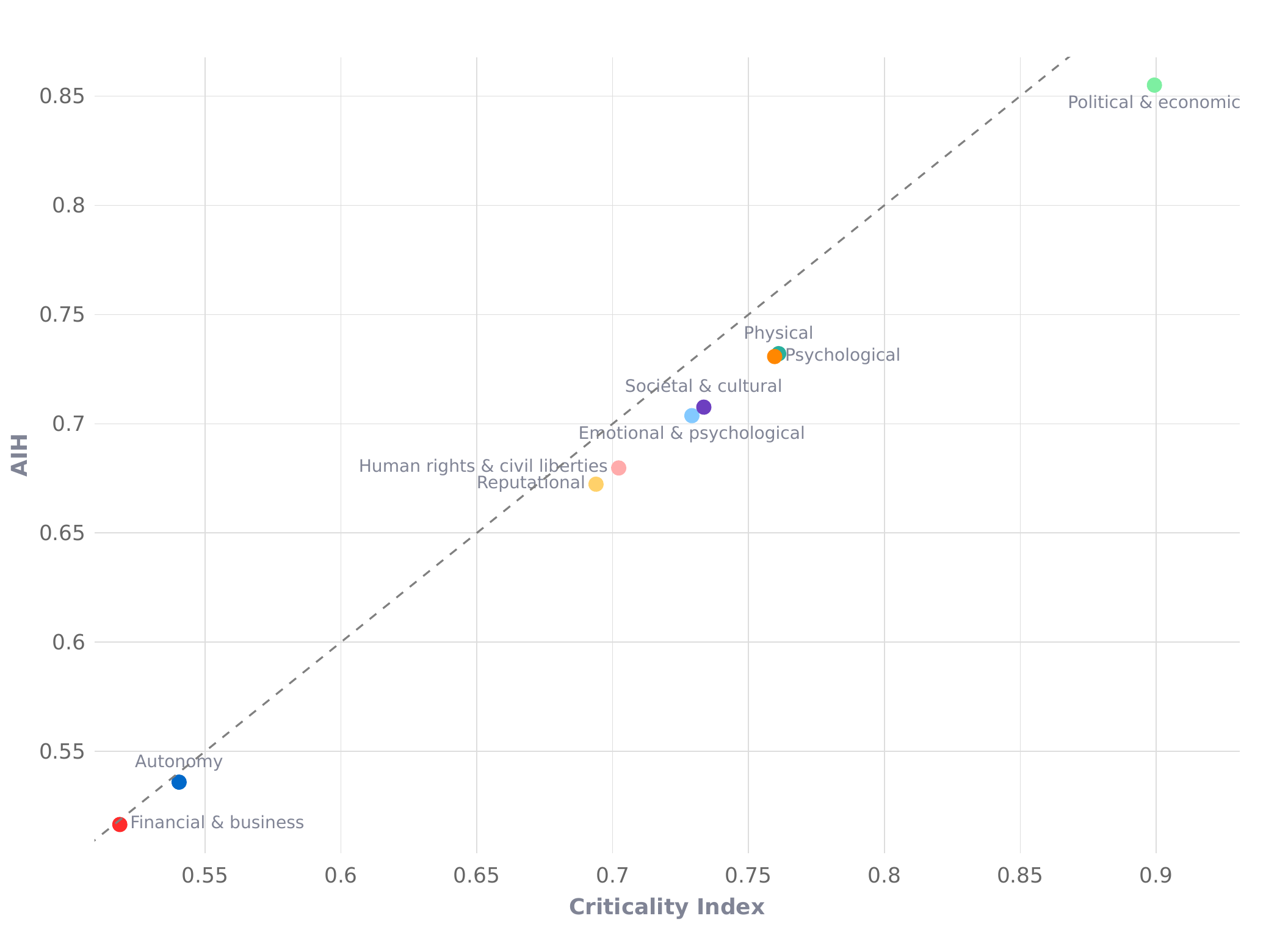}
\caption{Scatter plot comparing mean the \textit{AIH} for different harm categories using $\textit{AIH}$ and the CI. The dashed line indicates the 45-degree line.}
\label{scatter1}
\end{figure*}

These results are confirmed by the calculation of the CI as presented in Table \ref{tab:gini-criticality-index}. Its highest value, equal to $0.85$, is reached for the \textit{Political \& Economic} harm category, whereas its lowest values are attained for the \textit{Financial \& Business} category: $0.51$ and  the \textit{Autonomy} category: $0.53$. The slight differences between the \textit{AIH} and CI values reflect their underlying sensitivities: the \textit{AIH} metric emphasizes inequality at the extremes, while the CI captures average severity rank. This suggests that categories like \textit{Political \& Economic} exhibit sharper harm concentration, whereas categories with more balanced distributions, like \textit{Financial \& Business}, appear less critical under the \textit{AIH} metric.

Figure \ref{scatter1} compares the values of $\textit{AIH}$ against the CI. It reveals a clear positive trend (see Appendix~\ref{lsec:appendix-plotshsa} for the subcategories trend), indicating that categories with higher inequality under one metric tend to exhibit similar behavior under the other metric. From both metrics, for example, \textit{Financial \& Business} and \textit{Political \& Economic} have the lowest and highest $\textit{AIH}$ and CIs respectively.

\subsection{Sensitivity analysis}\label{sec:sensitivity-analysis}
The values of the $\textit{AIH}$ metric depend on the underlying severity values, and the results in Section~\ref{subs: hsa} have been calculated based on the order and on the values in Table~\ref{tab:stakeholders_severity}. In this section, we deviate from this initial choice and explore the variations in the results.

\subsubsection{Boundary Analysis}
In this section we calculate the maximum and minimum values of the $\textit{AIH}$ metric, under any possible mapping of severity ordering to the different stakeholder groups. These values reveal the bounds of the risk assessment for each category and potential limits of inequality under extreme conditions.

To calculate the maximum (respectively, minimum) $\textit{AIH}$ values, we proceed as follows: we sort stakeholders based on their frequency in each category in ascending (respectively, descending) order, and assign to them ascending ordering of the severity values. Then, the Lorenz curves and $\textit{AIH}$ metric are calculated (see Table \ref{tab:worst_best_gini} for all categories) for both extreme cases and all categories.

The combined Table \ref{tab:worst_best_gini} of best- and worst-case $\textit{AIH}$ coefficients provides insights into the range of inequality for different harm categories for the $\textit{AIH}$ metric. Categories such as \textit{Physical} which present a large $\textit{AIH}$ coefficient range, and \textit{Political \& economic} which shows, for example, a coefficient of $0.86$ in the worst case and $0.13$ in the best case display the largest disparity between the two scenarios (see Appendix \ref{lsec:appendix-plots} Figure~\ref{gba}), suggesting that harm inequality in this category is highly sensitive to how severity is distributed, confirming the findings of our previous experiments. On the other hand, the \textit{Financial \& Business} category ($\textit{AIH}$: $0.67$ worst, $0.35$ best) shows the most narrow range, reflecting a more balanced harm distribution regardless of the scenario, reflecting, again, the results for harm concentration in the previous section.

\begin{table}[htb!]
\centering
\footnotesize
\caption{Combined Table: Best case vs Worst case $\textit{AIH}$ for various Harm Categories for $\textit{AIH}$.}
\begin{tabular}{|p{3.5cm}|p{1.5cm}|p{1.5cm}|}
\hline
\textbf{Category} & \textbf{Best Case $\textit{AIH}$} & \textbf{Worst Case $\textit{AIH}$} \\ \hline
Autonomy & 0.19 & 0.80 \\ \hline
Physical & 0.13 & 0.86 \\ \hline
Psychological & 0.19 & 0.80 \\ \hline
Reputational & 0.23 & 0.76 \\ \hline
Financial \& business & 0.35 & 0.67 \\ \hline
Human rights \& civil liberties & 0.21 & 0.78 \\ \hline
Societal \& cultural & 0.24 & 0.76 \\ \hline
Political \& economic & 0.13 & 0.86 \\ \hline
Emotional \& psychological & 0.18 & 0.81 \\ \hline
\end{tabular}
\label{tab:worst_best_gini}
\end{table}

\subsubsection{Variations in ordinal values of severity}
In this section we study variations in the \textit{ordinal} values of severity. Specifically, we start from the initial severity value ordering (Table~\ref{tab:stakeholders_severity}) and conduct a set of experiments where we change/permutate the order of stakeholders; for example, a permutation can be to reverse the order of \textit{Business} (from severity 3 to severity 4) and \textit{Investors}  (from severity 4 to severity 3), which means that we change the ordering.

\textbf{Experiment:} For the first set of scenarios (\textit{1 Permutation}) we first do a single permutation (vs. the original scenario in Table~\ref{tab:stakeholders_severity}) for a pair of consequent severity levels. We generate 20 scenarios, each of them randomly selecting a pair of severity levels to permute. The second set of scenarios (\textit{2 Permutations}) follows the same setup, with the only difference that we do two random severity permutations instead on one. And, similarly for the \textit{5 Permutations} set of scenarios.

\begin{figure*}
\centering
\includegraphics[width=0.9\textwidth]{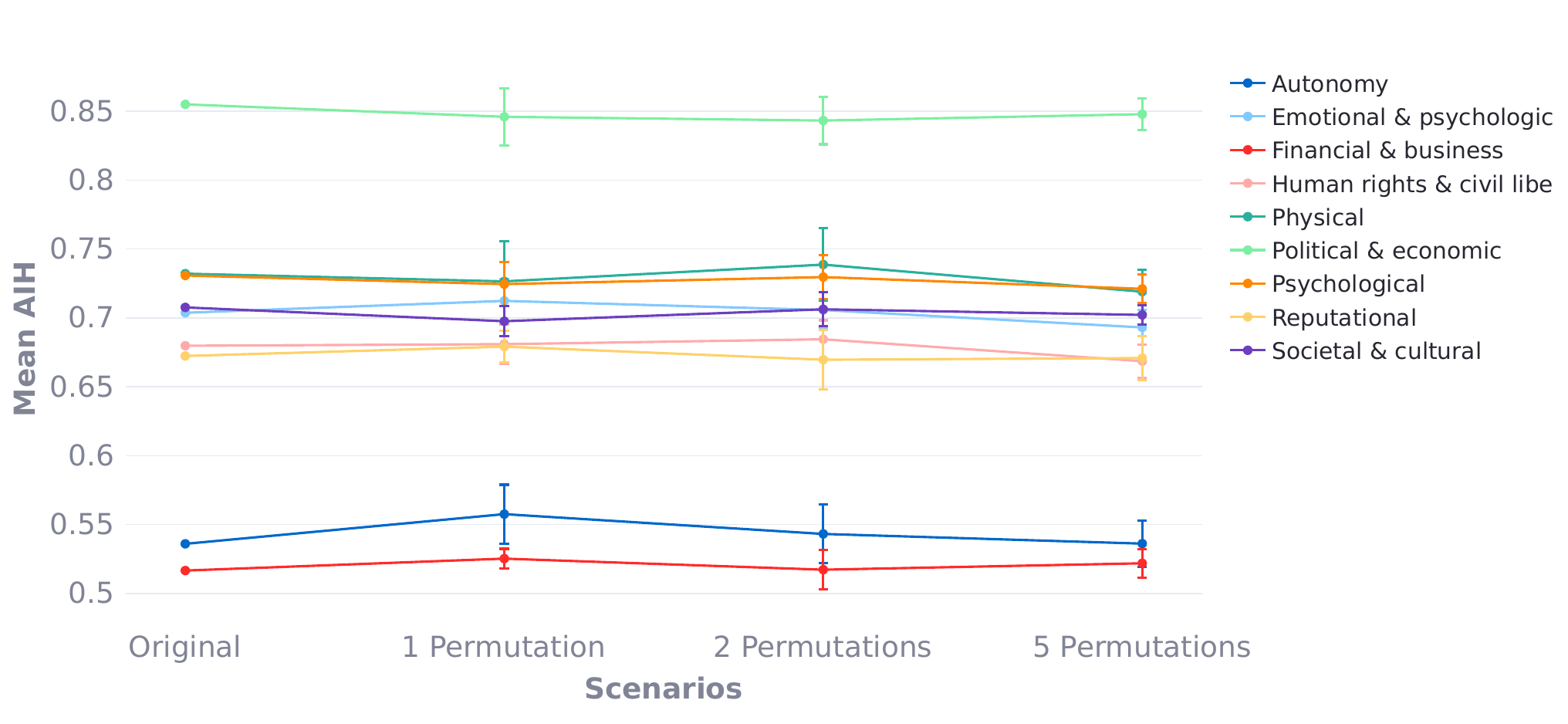}
\caption{Mean $\textit{AIH}$ per scenario for variations in ordinal values of severity for $\textit{AIH}$.}
\label{meangperscenario1}
\end{figure*}

\begin{figure*}
\centering
\includegraphics[width=0.6\textwidth]{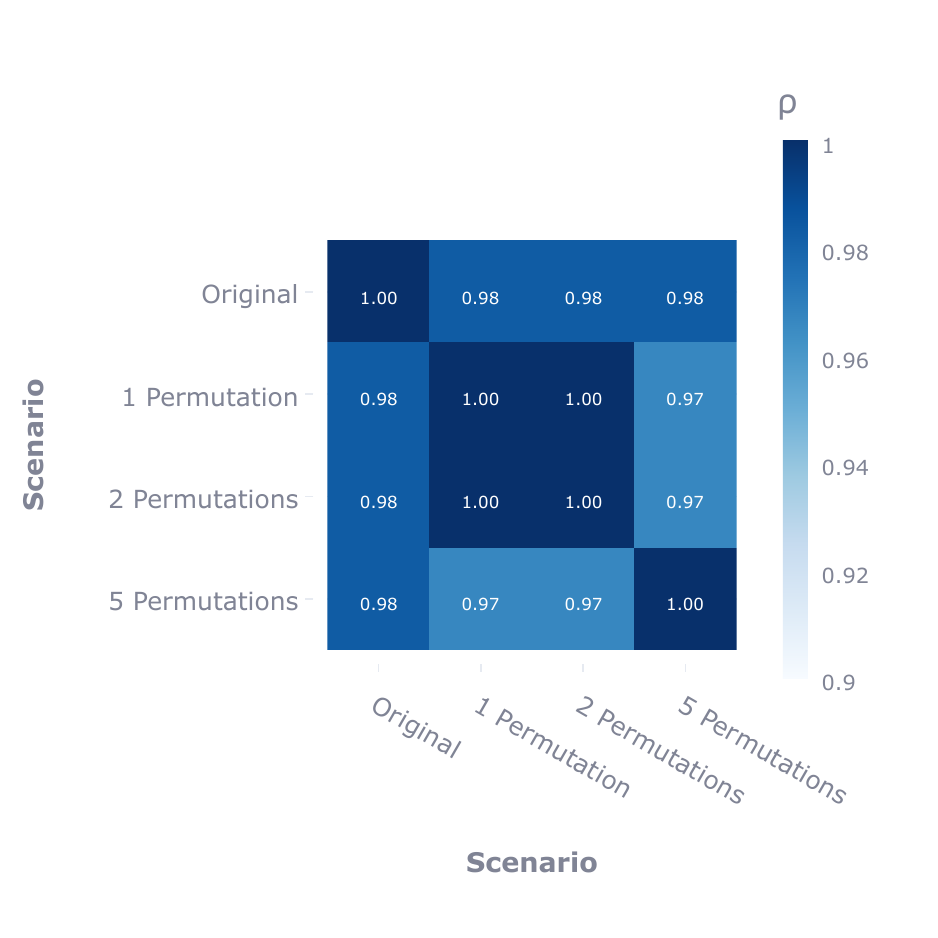}
\caption{Spearman’s rank correlation ($\rho$) between scenarios of ordinal‐severity perturbations for $\textit{AIH}$.}
\label{fig:spearman_ord}
\end{figure*}

Figure \ref{meangperscenario1} shows the mean values of the $\textit{AIH}$ metric across the three sets of scenarios (see Appendix~\ref{lsec:appendix-plots} Figures \ref{fig:drill_boxplot_pe}, \ref{fig:drill_scatter_pe} for an even more detailed analysis of the harm subcategories of the harm category that tends to have the highest $\textit{AIH}$; Political \& Economic). The results indicate that harm category rankings remain stable across permutations, indicating that the concentration of harm is an inherent property of the collected data and relatively robust to small variations in the selected severity values. Category rankings are highly consistent, with minimal variation and strict dominance even under multiple perturbations, reinforcing its robustness.

This is further confirmed in Figure~\ref{fig:spearman_ord}, which presents the pairwise Spearman rank correlations ($\rho$) between the harm category rankings produced by the $\textit{AIH}$ metric under different ordinal severity permutations. Each cell reflects the rank similarity between two scenarios: the \textit{Original} ordering and versions with 1, 2, or 5 random permutations in the severity order. As expected, the diagonal values are $\rho = 1.00$, since each scenario is perfectly correlated with itself. However, the off-diagonal values are what validate the robustness of our approach: they remain consistently high across all permutations; between $0.97$ and $1.00$.

This demonstrates that even when the severity order is altered randomly up to five times, the relative ranking of harm categories remains nearly unchanged. For instance, the correlations between the original and all perturbed scenarios are all greater than or equal to $0.98$, which implies a very strong agreement in harm prioritization regardless of minor perturbations.

These results confirm that the concentration patterns captured by $\textit{AIH}$ are not fragile or overly dependent on a specific severity ordering. Furthermore, they show that our harm-ranking method which uses $\textit{AIH}$ is stable under realistic uncertainty or disagreement in how severity levels are ranked across stakeholder groups.

\subsubsection{Variations by Random Annotation Removal}

Real‐world incident data are often incomplete or continuously growing, raising the question: how sensitive are our concentration metrics to the size of the underlying annotation set? To address this, we simulate progressively “pruned” versions of the AIAAIC dataset by randomly dropping fixed fractions of the original annotations and then recompute the $\textit{AIH}$ exactly as before. This tests whether our harm‐category rankings remain stable when only a subset of stakeholder–severity pairs is available.

\textbf{Experiment:} Starting from the full set of 816 annotations, we define four removal scenarios - 10\%, 20\%, 50\%, and 80\% random deletion - plus the baseline (0\% removed). For each, we investigate how $\textit{AIH}$ changes for each harm category (or subcategory).

\begin{figure*}
\centering
\includegraphics[width=0.9\textwidth]{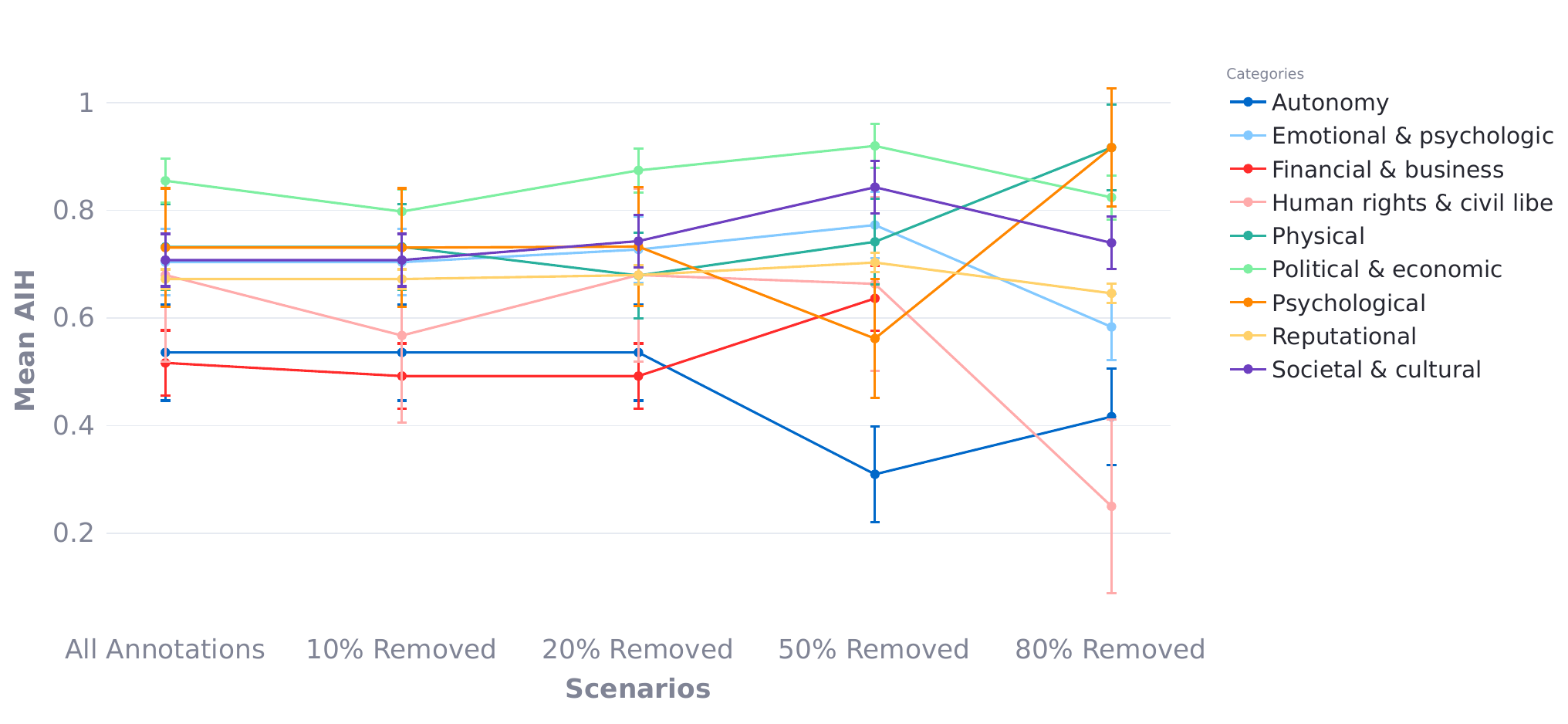}
\caption{Mean $\textit{AIH}$ for each harm category under each progressive random annotation removal. Points mark the mean $\textit{AIH}$; vertical bars show 95\% confidence intervals.}
\label{fig:removal-trend-gini}
\end{figure*}

Figure~\ref{fig:removal-trend-gini} shows that up to 20\% removal, every category’s $\textit{AIH}$ shifts by only a few hundredths and the ranking - \emph{Political \& Economic} highest, \emph{Financial \& Business} and \emph{Autonomy} lowest - remains stable. At 50\% removal, curves begin to spread slightly (e.g.\ the \emph{Autonomy} line dips more) while the mean $\textit{AIH}$ of \emph{Reputational} harm category seems to get larger, but no category overtakes another to a great extent. Even under the extreme 80\% deletion, although some $\textit{AIH}$ values drop, such as \emph{Reputational}, indicating a more uniform harm spread since random deletion disproportionately affects high‐severity annotations, the identity of the “most” and “least” concentrated categories stays intact see Appendix~\ref{lsec:appendix-plots} 
 Figure \ref{fig:boxplotspe} for the boxplots of the harm subcategories of the harm category that tends to have the highest concentration; Political \& Economic).

These observations imply that random removal of annotations tends to dilute extreme severity concentrations - hence the mild downward drift in $\textit{AIH}$, but does not reorder the harm priorities. In practice, this stability ensures that policymakers can confidently prioritize the same harm categories even when working with incomplete or evolving incident datasets.

\section{On the Suitability of Incident Datasets} 
\label{sec:suitability}
To analyze AI-related harms from an external perspective - emphasizing the experiences and impacts on various stakeholder groups - it is preferable to utilize datasets that document the affected parties, as this enhances the granularity and stakeholder-orientation of the analysis. However, the framework can still be applied when only harm categories and severity levels are available. While several AI incident and risk databases exist, many fall short in providing the necessary granularity regarding stakeholder information.

The MITRE ATT\&CK framework \cite{strom2018mitre} and the MITRE ATLAS framework \cite{MITREATLAS} are widely recognized resources for understanding security-related aspects of AI systems. MITRE ATT\&CK focuses on adversarial tactics, techniques, and procedures (TTPs) used in cybersecurity threats, while MITRE ATLAS extends this by offering a comprehensive taxonomy and case studies on methods for intentionally attacking or exploiting AI systems. While these frameworks provide invaluable insights for threat modeling and adversarial robustness, their primary scope remains firmly rooted in cybersecurity risks. Crucially for our purposes, they do not address the broader spectrum of societal AI harms, nor do they provide information about the specific stakeholder groups affected by these harms. However, for a user whose need is to investigate the harm concentration for specifically the cybersecurity sector, our framework can be very useful to apply.

Shifting away from cybersecurity and towards the documentation of real-world AI failures, the AI Incident Database (AIID) \cite{mcgregor2021preventing}, maintained by the Partnership on AI, is a prominent initiative that collects real-world reports of AI-related incidents across diverse domains. It focuses on documenting cases where AI systems have caused or nearly caused harm, regardless of intent, making it a rich source of qualitative information about system-level failures. However, despite the breadth of incidents it captures, the AIID does not systematically record detailed stakeholder information, which limits stakeholder-specific insights. Nonetheless, the dataset can still be used in our framework with severity data aggregated per harm category.

Building upon the AIID dataset, the MIT AI Risk Repository \cite{slattery2024airiskrepositorycomprehensive} does not constitute a repository of incidents itself, but rather serves as a structured framework for classifying AI risks. It offers two taxonomies - causal and domain-based - that are designed to organize known AI risk types by intent, responsible entity, and sectoral impact. While this taxonomy has proven useful for consistent classification and has been applied in automated analyses of AIID records (e.g., through the AI Incident Tracker project), the repository does not include incident data on its own, nor does it incorporate stakeholder-level annotations. Thus, while the MIT framework contributes important conceptual infrastructure for risk assessment, it cannot independently support stakeholder-oriented harm analysis without being combined with external incident data and additional stakeholder tagging.

Complementary to AIID, but focusing at a more technical and operational level, the AI Vulnerability Database \cite{AVID2024} provides a structured repository of both intentional attacks and unintentional failures of AI systems. Distinguishing itself from AIID, AVID offers reproducible technical details, down to specific models and datasets, and organizes failure modes using a fine-grained operational taxonomy. Its primary objective is to support AI developers and auditors in identifying vulnerabilities and system weaknesses. Yet, despite its technical depth, AVID, like the aforementioned resources, does not document or categorize the specific stakeholders impacted by these vulnerabilities. It is fundamentally designed for technical risk evaluation rather than stakeholder-oriented harm assessment, which remains the focus of this work.

The Database of AI Litigation (DAIL) \cite{DAIL2024}, curated by the George Washington University Law School's Ethical Tech Initiative, serves as a repository for ongoing and completed litigation involving artificial intelligence, including machine learning. It encompasses a broad spectrum of cases, from algorithms used in hiring and credit decisions to liability issues involving autonomous vehicles. While DAIL provides valuable insights into the legal challenges associated with AI technologies, it does not systematically document the specific stakeholders affected by each case. The database focuses on the legal aspects of AI incidents, such as the nature of the litigation and the legal questions involved, rather than on the detailed impacts on various stakeholder groups, which again, makes it less suitable for analyses aimed at understanding the differential impacts of AI across diverse populations.

A notable exception in this landscape is the OECD AI Incidents Monitor (AIM) \cite{OECD2024Incidents}, which indeed provides information about the stakeholders impacted by AI incidents. The OECD AIM constitutes an important initiative for documenting AI-related incidents globally, utilizing a combination of international news sources and automated machine learning (ML) models to identify and categorize events. Its framework supports the annotation of harm types, severity, and entities harmed, including stakeholders, thus aligning in principle with the needs of external, stakeholder-focused harm analysis. However, those "severity" in this dataset's case involves the type of severity, not severity levels, leaving users to infer or assign their own ordinal scales. The methodology employed by the OECD AIM to identify stakeholders relies primarily on automated information extraction techniques applied to publicly available sources, predominantly news articles. While this approach enables the rapid collection of large-scale data, it poses significant challenges for our purposes. The reliance on unstructured textual descriptions, combined with automated classification using large language models (LLMs), often results in stakeholder information being inconsistently captured, especially in incidents where the harm context is complex or underreported. The OECD AIM provides valuable aggregated mappings of harms and affected entities but it lacks expert-driven stakeholder annotations, who are replaced with LLMs, but our framework applying $\textit{AIH}$ remains perfectly applicable using the incident or category-level severity ratings it provides.

For these reasons, in this study, we adopt the AIAAIC dataset \cite{aiaaic}, built on the taxonomy of AI, algorithmic and automation harms \cite{abercrombie2024collaborative}, which was explicitly created to capture real-world AI incidents. Crucially, the AIAAIC dataset that we analyze includes detailed and validated information on the stakeholders affected by each incident and their harm category and subcategory, after the annotation process of human experts within the activities of the initiative. This granular, consensus-based approach enables us to analyze harm severity and its distribution across stakeholder groups with the necessary reliability and precision - a prerequisite for developing risk assessment methodologies that prioritize external impacts over internal compliance metrics.

We note that the tables and structures we present for the AIAAIC dataset in this work should be regarded as indicative of the form that any suitable dataset should follow in order to be compatible with our proposed framework. Specifically, our framework is dataset-agnostic: any dataset that contains comparable information - namely, a structured representation of incidents including affected stakeholders, harm categories (and subcategories) and severity annotations - can be seamlessly integrated into our framework, and we additionally provide the dataset we preprocessed after it was annotated as a benchmark.  For example, incident records from databases like the MIT AI Incident Tracker dataset and the OECD AI Incidents Monitor could also be good fits for this structure and support similar harm severity assessments. We proceed by explaining how the former could also be used in our framework.

As for the MIT AI Incident Tracker dataset, its structure differs from that of the AIAAIC dataset in several important ways. In AIAAIC, multiple annotations may exist for a single incident, each documenting a unique combination of stakeholder group, harm category, and an associated severity value, as can be seen in Table \ref{tab:aiaaic-example}. In contrast, the MIT dataset presents each row as a distinct incident, with severity assessments - referred to as “ratings” - assigned directly to various harm categories, without specifying which stakeholder groups are affected.

To adapt this structure to our framework, we consider each row in the MIT dataset as a single annotation, where the number of annotations is equal to the number of incidents. For each harm category, multiple ratings across different incidents are aggregated. These ratings functionally play the role of severity values in our framework. Thus, for each harm category, we compute the frequency of incidents receiving a given severity rating, resulting in a (harm category, severity, frequency) table that is structurally analogous to the stakeholder-based tables used for AIAAIC. This adjustment allows us to apply the same metric to quantify harm concentration, despite the absence of explicit stakeholder information.

For example, in Table~\ref{tab:mit-example}, each row represents one such (harm category, rating, frequency) triplet. These records serve as input to our framework in a structure that parallels the stakeholder-based formulation used for AIAAIC, enabling the calculation of $\textit{AIH}$ to assess harm concentration.

\newcommand{\thickvrule}{\vrule width 1.5pt}

\setlength{\arrayrulewidth}{0.6pt}

\begin{table}[ht]
  \centering
  \caption{\textbf{AIAAIC dataset:} Example of essential values for the AIAAIC dataset needed to apply $\textit{AI Harmonics}$ framework. Each row represents a different annotation, while many annotations might represent one incident.}
  \label{tab:aiaaic-example}
  \begin{tabular}{|l|l!{\thickvrule}c|c!{\thickvrule}}
    \hline
    \textbf{Harm Category} & \textbf{Stakeholder} & \textbf{Severity} & \textbf{Freq.} \\
    \hline
    A & a & 1 & 10 \\
    A & b & 2 & 15 \\
    B & c & 3 & 29 \\
    B & a & 1 & 25 \\
    C & a & 1 & 21 \\
    C & b & 2 & 11 \\
    \hline
    … & … & … & … \\
    \hline
  \end{tabular}
\end{table}

\begin{table}[ht]
  \centering
\captionsetup{justification=raggedright,singlelinecheck=false}
  \caption{\textbf{MIT (AI Incident Tracker) dataset:} Example of essential values needed to apply $\mathit{AI Harmonics}$ framework. Each row represents a different incident, which is considered a separate annotation.}
  \label{tab:mit-example}
  \begin{threeparttable}
    \begin{tabular}{|l!{\thickvrule}c|c!{\thickvrule}}
      \hline
      \textbf{Harm Category} 
        & \textbf{Rating}\tnote{a} 
        & \textbf{Freq.} \\
      \hline
      A & 1 & 10 \\
      A & 2 & 15 \\
      B & 3 & 29 \\
      B & 1 & 25 \\
      C & 1 & 21 \\
      C & 2 & 11 \\
      \hline
      … & … & … \\
      \hline
    \end{tabular}
    \begin{tablenotes}[flushleft]
      \footnotesize
      \item[a] Exactly respective to "Severity".
    \end{tablenotes}
  \end{threeparttable}
\end{table}

\begin{table}[ht]
  \centering
  \captionsetup{justification=raggedright,singlelinecheck=false}
  \caption{\textbf{OECD AI Incidents and Hazards Monitor dataset:} 
    Example of essential values for the OECD dataset needed to apply
    $\mathit{AI Harmonics}$ framework. Each row represents a different
    annotation, while many annotations might represent one incident.}
  \label{tab:oecd-example}
  \begin{minipage}{\textwidth}
    \centering
    \begin{threeparttable}
      \begin{tabular}{|l|l!{\thickvrule}c|c!{\thickvrule}}
        \hline
        \textbf{Harm Type}\tnote{a}
          & \textbf{Stakeholder}
          & \textbf{Severity}\tnote{b}
          & \textbf{Freq.} \\
        \hline
        A & a & 1 & 10 \\
        A & b & 2 & 15 \\
        B & c & 3 & 29 \\
        B & a & 1 & 25 \\
        C & a & 1 & 21 \\
        C & b & 2 & 11 \\
        \hline
        … & … & … & … \\
        \hline
      \end{tabular}
      \begin{tablenotes}[flushleft]
        \footnotesize
        \item[a] Exactly respective to "Harm Category".
        \item[b] Assigned by the user or LLMs. Not already present in the dataset.
      \end{tablenotes}
    \end{threeparttable}
  \end{minipage}
\end{table}

This adaptation allows us to retain the core logic of our framework, mapping harm categories to severity levels and calculating harm concentration, while accommodating datasets that do not include explicit stakeholder - level annotations. In practice, this means that while the granularity and interpretation of severity may differ (individual vs. aggregated stakeholder impact), the analytical pipeline remains robust. Moreover, this flexibility highlights one of the strengths of our framework: its ability to integrate heterogeneous sources of harm data for unified, comparative analysis.

To adapt the OECD AIM data to our framework, we treat each row - each combination of Harm Type (which is directly analogous to our Harm Category), harmed entity (Stakeholder), and source incident - as a separate annotation, much as we did with the MIT AI Incident Tracker dataset. Since the OECD Monitor does not publish any explicit "severity levels," users must supply them either manually (e.g.\ via expert judgment) or via LLM‐assisted annotation pipelines; these assigned values then serve exactly the same role as our stakeholder‐severity rankings in Table \ref{tab:stakeholders_severity}. Once a severity score is attached to each (Harm Type, Stakeholder) pair, we aggregate across all rows to compute a triplet (Harm Type, Stakeholder, Frequency), exactly according to how we treat AIAAIC dataset in Table \ref{tab:aiaaic-example}.

Because Harm Type in the OECD schema is semantically identical to our notion of Harm Category, no additional mapping is needed: all rows are simply grouped by Harm Type, tally the frequency of each Stakeholder at each assigned severity level, and feed those counts into the $\mathrm{AIH}$ or CI calculations.

Although the automated stakeholder extraction in the OECD Monitor offers broad coverage, it may misclassify or miss less‐prominent harmed entities; likewise, severity assignments (whether human‐ or LLM‐generated) introduce additional noise. However, our sensitivity analyses (Section \ref{sec:sensitivity-analysis}) demonstrate that $\mathrm{AIH}$ remains robust under such perturbations, making the OECD dataset a fully viable candidate for harm‐concentration studies alongside AIAAIC and MIT AI Incidents Tracker.

\section{Conclusions} \label{sec:conclusion}

In this work we have introduced AI Harmonics, a fully stakeholder-centric, harm-focused paradigm for AI risk assessment that operates even when precise numerical severities are unavailable. Central to our approach is a novel ordinal concentration metric, ($\textit{AIH}$), which measures how unevenly harms are distributed across ranked severity levels. Because it requires only a total ordering of severity judgments, $\textit{AIH}$ unlocks rigorous prioritization in domains where quantitative loss estimates are scarce or unreliable. At the same time, $\textit{AIH}$ seamlessly collapses to the classic numerical Gini when actual severity values are provided, offering a unified framework that spans purely ordinal and fully quantitative scenarios.

We also contribute to a stakeholder-embedding methodology: by mapping incident annotations from the AIAAIC repository onto a shared harm taxonomy and nine stakeholder groups, which we use as a benchmark, we give voice to those most affected by AI failures - citizens, workers, vulnerable populations, and beyond - rather than relying solely on provider-centric compliance checks. Additionally, we validate the $\textit{AIH}$ metric against the established Criticality Index (CI), confirming that our approach captures harm inequality under ordinal inputs.

Finally, we establish the robustness and adaptiveness of AI Harmonics through extensive sensitivity analyses. We show that category rankings remain effectively unchanged under random permutations of severity orderings and under the removal of up to 80 \% of annotations, assuring users that prioritization is driven by genuine concentration patterns rather than quirks of any single dataset.

Empirical application to the AIAAIC dataset highlights Political \& Economic harms as the most sharply concentrated, demanding urgent attention, while Financial \& Business and Autonomy harms exhibit more uniform distributions. Our dataset-agnostic pipeline can be plugged into any incident repository with ordered severity labels (or extended numerical scores), such as  the OECD AI Incidents Monitor or the MIT AI Incident tracker,  and we provide an open-source implementation alongside an interactive dashboard for exploring Lorenz curves, the $\textit{AIH}$ metric, and sensitivity scenarios. By combining stakeholder perspectives, ordinal scales, and extensions to numeric measures, AI Harmonics offers a flexible, reproducible tool for policymakers and practitioners to pinpoint and mitigate the most critical AI-driven harms.

\section*{CRediT authorship contribution statement}
\textbf{Sofia Vei}: Data curation, Formal analysis, Methodology, Software, Validation, Visualization, Writing – original draft. \textbf{Paolo Giudici}: Conceptualization, Methodology, Writing – Review \& Editing, Supervision. \textbf{Pavlos Sermpezis}: Data Curation, Writing – Original Draft. \textbf{Athena Vakali}: Writing – Review \& Editing, Supervision. \textbf{Adelaide E.~Bernardelli}: Writing – original draft.

\section*{Declaration of Competing Interest}
The authors declare that they have no known competing financial interests or personal relationships that could have appeared to influence the work reported in this paper.

\section*{Acknowledgments}
We wish to thank the AIAAIC initiative for providing access to the anonymized, expert-annotated dataset, and in particular Djalel Benbouzid for supplying the raw annotations that we subsequently preprocessed.\\
The work is the result of a close collaboration among the authors. The work of Paolo Giudici has been funded by the European Union – NextGenerationEU, in the framework of the GRINS – Growing Resilient, INclusive and Sustainable (GRINS PE00000018). The work of A. E. Bernardelli has been conducted during and with the support of the Italian national inter-university PhD course in Sustainable Development and Climate change (link: www.phd-sdc.it), Cycle XL, with the support of a scholarship co-financed under Ministerial Decree no. 630 of 28th March 2024 (Investment 3.3 – Innovative PhDs), based on the National Recovery and Resilience Plan (NRRP) - funded by the European Union - NextGenerationEU - Mission 4 "Education and Research", Component 2 "From Research to Business", and by Noto Sondaggi Srl.

\section*{Code \& Data Availability}
An interactive demonstration of the AI Harmonics framework, featuring live visualizations and user-driven exploration of harm concentration metrics, is publicly available \cite{aih-webapp}.  
For full transparency and reproducibility, the complete source code, detailed installation instructions, and the pre-processed AIAAIC incident dataset used as a benchmark in this study can be found in our GitHub repository \cite{aih-github}.  
We invite researchers and practitioners to clone the repository, reproduce our analyses, and adapt the AI Harmonics pipeline to their own incident datasets.

\bibliography{paper}
\bibliographystyle{elsarticle-num} 

\clearpage
\FloatBarrier
\appendix
\section{Appendix: $\textit{AIH}$ vs $\textit{CI}$} \label{lsec:vs}

\begin{center}
    \begin{threeparttable}
  \captionof{table}{Pros and Cons of Gini Index vs.\ Criticality Index for AI Harm Prioritization.}
  \label{tab:pros_cons}
  \small
  \begin{tabularx}{\linewidth}{Xcc}
    \toprule
      & \textbf{Gini Index} & \textbf{Criticality Index} \\
    \midrule
    \multicolumn{3}{l}{\emph{Pros}}\\
    Captures inequality/fairness across stakeholders\tnote{1}  
      & \cmark &  \\
    Supports pure ordinal severity scales         
      & \cmark & \cmark \\
      Supports pure ordinal severity scales         
        & \cmark & \cmark \\
      Emphasizes worst‐case (extreme) harms         
        &  & \cmark \\
      Intuitive (area‐under‐curve) interpretation   
        & \cmark &  \\
      Integrates with broader fairness metrics\tnote{2}      
        & \cmark &  \\
      Computational simplicity                      
        & \cmark & \cmark \\
      Well‐established and standardized\tnote{3}              
        & \cmark &  \\
      \midrule
      \multicolumn{3}{l}{\emph{Cons}}\\
      Assumes equal‐interval severities if numeric\tnote{4}  
        & \cmark &  \\
      Overemphasizes extremes (may miss moderate but widespread harms) 
        &  & \cmark \\
      No widely adopted standard or software implementation     
        &  & \cmark \\
      Less intuitive (average‐rank based)           
        &  & \cmark \\
      \bottomrule
    \end{tabularx}
    \begin{tablenotes}
      \footnotesize
      \item[1] Gini measures the full shape of the distribution against perfect equality, whereas CI only reports the average rank without an equality baseline.
      \item[2] Gini is part of the standard inequality‐measurement toolkit (e.g.\ income, resource allocation), making it easy to combine with other fairness metrics; CI is more specialized to risk ranking.
      \item[3] Gini is universally defined in textbooks, statistical software, and policy reports; CI has no single standard reference or widespread implementation.
      \item[4] In our methodology we use the ordinal variant ($\textit{AIH}$), so no equal‐interval assumption is ever introduced.
  \end{tablenotes}
  \end{threeparttable}

\end{center}

Table~\ref{tab:pros_cons} summarizes the respective strengths and weaknesses of $\textit{AIH}$ and CI. Note in particular how our framework overcomes the main con of the classic Gini - namely, its reliance on equal‐interval severity values - since $\textit{AIH}$ never uses the actual numeric spacing. Overall, the methodology maintains full compatibility with ordinal data, making it a powerful, general‐purpose tool for stakeholder‐centric AI harm prioritization.

\FloatBarrier
\section{Appendix: Definitions of Stakeholder groups and Detailed list of Subcategories}\label{lsec:appendix-data}

\begin{table*}
\tiny
\centering
\caption{Stakeholder groups and their definitions}
\begin{tabular}{|p{4cm}|p{10cm}|}
\hline
\textbf{Stakeholder Group} & \textbf{Definition} \\
\hline
Artists/Content Creators & People inventing, producing or making creative and/or IP/copyright-protected products, services, or content. \\
\hline
Business & Companies and organizations impacted by AI in terms of operations, competition, regulatory compliance, or financial performance. \\
\hline
General public & Individuals or entities such as passers-by, local communities and other members of the general public indirectly harmed by a system, often without their knowledge. \\
\hline
Government/Public Sector (Service) & Including politicians, civil servants, and regulators. \\
\hline
Investors & Shareholders/investors in the developer and/or deployer. \\
\hline
Subjects & Individuals or entities such as patients, students, employees, delivery drivers, travellers, immigrants and animals harmed when targeted by a technology system over which they have little or no control or direct access to its data or outputs. \\
\hline
Users & Individuals or entities such as citizens, consumers, job and visa applicants harmed when using or consciously interacting with a technology system. \\
\hline
Vulnerable groups & Including women, children, disabled people, ethnic and religious minorities. \\
\hline
Workers & Employees and third-party contractors, vendors, gig workers and others tasked with training, managing and/or optimising data or information systems \\
\hline
\end{tabular}
\label{tab:stakeholders_definitions}
\end{table*}

\begin{table*}
\tiny
\centering
\caption{Categories and their respective subcategories of harms caused by AI systems}
\begin{tabular}{|p{4cm}|p{10cm}|}
\hline
\textbf{Harm Category} & \textbf{Harm Subcategories} \\
\hline
Autonomy & 
Autonomy/agency loss, Impersonation/identity theft, Personality loss, IP/copyright loss, Personality rights loss \\
\hline
Physical & 
Loss of life, Bodily injury, Property damage, Personal health deterioration \\
\hline
Psychological & 
Coercion/manipulation, Anxiety/distress, Sexualisation, Dehumanisation/objectification, Dignity loss, Alienation/isolation, Over-reliance, Addiction, Harassment/abuse/intimidation, Self-harm, Radicalisation \\
\hline
Reputational & 
Defamation/libel/slander, Loss of confidence/trust \\
\hline
Financial \& business & 
Financial/earnings loss, Confidentiality loss, Loss of productivity, Opportunity loss, Livelihood loss, Business operations/infrastructure damage \\
\hline
Human rights \& civil liberties & 
Discrimination, Benefits/entitlements loss, Loss of human rights and freedoms, Privacy loss, Loss/violation of human rights and freedoms, Loss of right to due process, Loss of right to liberty and security \\
\hline
Societal \& cultural & 
Public service delivery deterioration, Stereotyping, Information ecosystem degradation, Violence/armed conflict, Damage to public health, Societal destabilisation, Societal inequality, Job loss/losses, Loss of creativity/critical thinking, Cheating/plagiarism \\
\hline
Political \& economic & 
Political instability, Institutional trust loss, Critical infrastructure damage, Political manipulation, Economic instability, Electoral interference, Economic/political power \\
\hline
Emotional \& psychological & 
Anxiety/distress/depression, Intimidation, Radicalisation, Dignity loss, Dehumanisation/objectification, Sexualisation, Self-harm, Addiction, Over-reliance, Alienation/isolation, Coercion/manipulation \\
\hline
\end{tabular}
\label{tab:categories_subcategories}
\end{table*}

\clearpage
\section{Appendix: Additional experimental results for Harm Severity Analysis}\label{lsec:appendix-plotshsa}

\begin{figure*}
\centering
\includegraphics[width=0.8\textwidth]{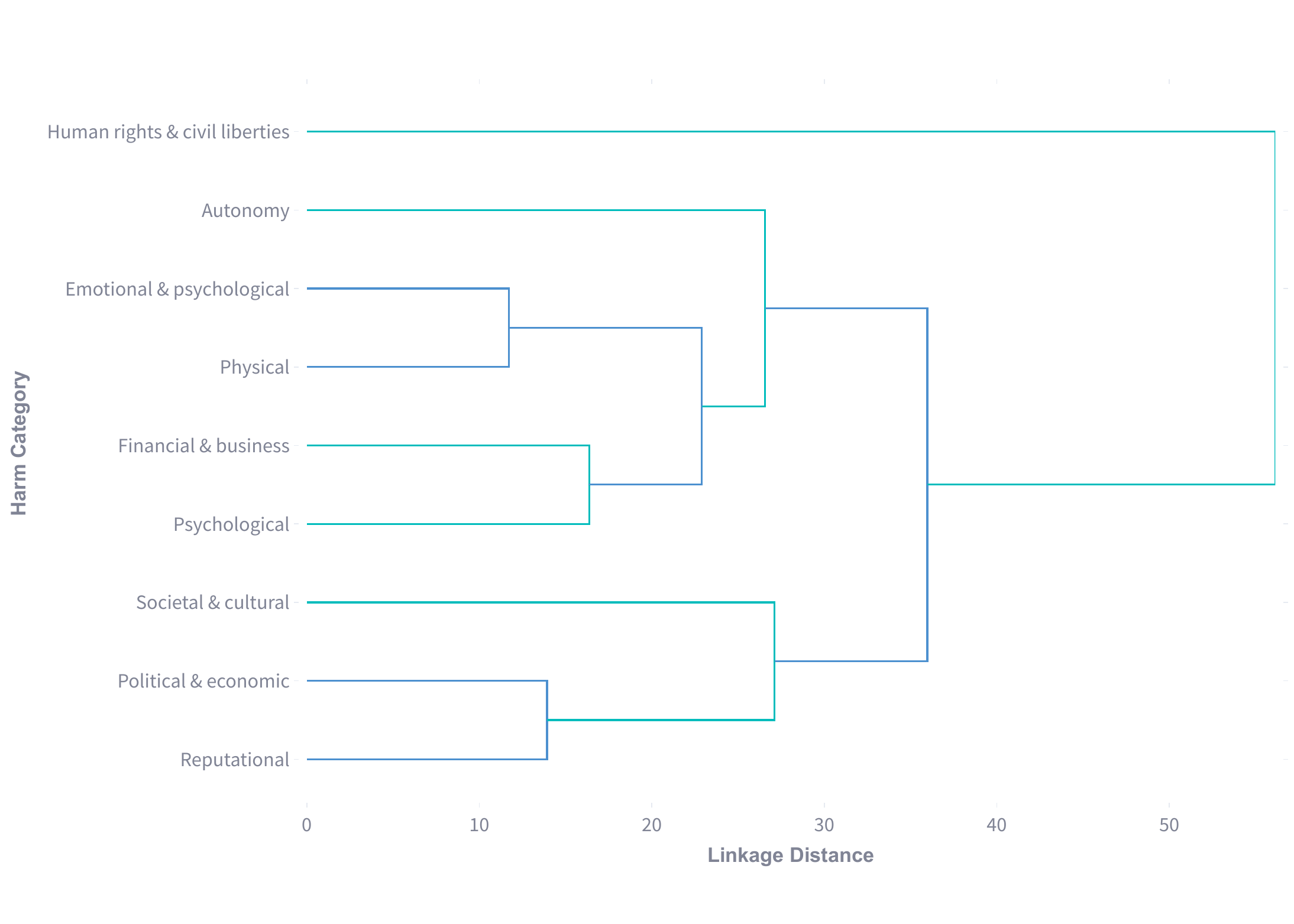}
\caption[Hierarchical Clustering of Harm Categories]{\textbf{Hierarchical clustering of harm categories by stakeholder impact.}  
    Linkage distance on the horizontal axis measures dissimilarity in the distribution of incident counts across stakeholder groups.  
    Categories that merge at low distance share very similar stakeholder‐impact profiles, while those that join only at high distance differ markedly.}
\label{fig:category_dendrogram}
\end{figure*}

\noindent
Figure~\ref{fig:category_dendrogram} refines the picture sketched by the heatmap, grouping harm categories according to how similarly they affect the nine stakeholder groups. The dendrogram is generated by computing pairwise Euclidean distances between each category’s vector of stakeholder incident counts, then applying Ward’s linkage method to those distances to produce the hierarchical clustering shown. At the lowest linkage distance, \emph{Emotional \& psychological} and \emph{Physical} harms merge first, indicating these two categories tend to burden the same mix of stakeholders to a similar degree.  These join with \emph{Autonomy}, suggesting that loss of agency often co-occurs with emotional distress and physical harms among the same parties. In parallel, \emph{Financial \& business} and \emph{Psychological} harms cluster together, reflecting a shared profile of impact (e.g.\ workers and vulnerable groups are relatively more affected).

At a higher level, \emph{Reputational} and \emph{Societal \& cultural} harms form a distinct branch, implying that reputational damage and broad societal‐cultural disruptions - while both widespread - nevertheless differ in which stakeholder groups they most deeply affect.  Finally, \emph{Human rights \& civil liberties} stands apart until the very last merge, underscoring its unique pattern: it concentrates most heavily on \emph{users} and \emph{vulnerable groups} (as the heatmap showed), and much less so on investors or subjects, making its distribution strikingly distinct from all other categories.

Together, the dendrogram and the preceding heatmap reveal both the magnitude and the structure of AI-related harms: the heatmap highlights which stakeholders bear the greatest raw counts in each category, while the clustering dendrogram uncovers natural groupings of harm types that share similar "victim profiles".  These insights can guide policy and mitigation design by identifying clusters of harms that may be addressed with common interventions, as well as pinpointing truly unique categories, like human-rights violations, that require targeted remedies.

\begin{figure*}
\centering
\includegraphics[width=0.7\textwidth]{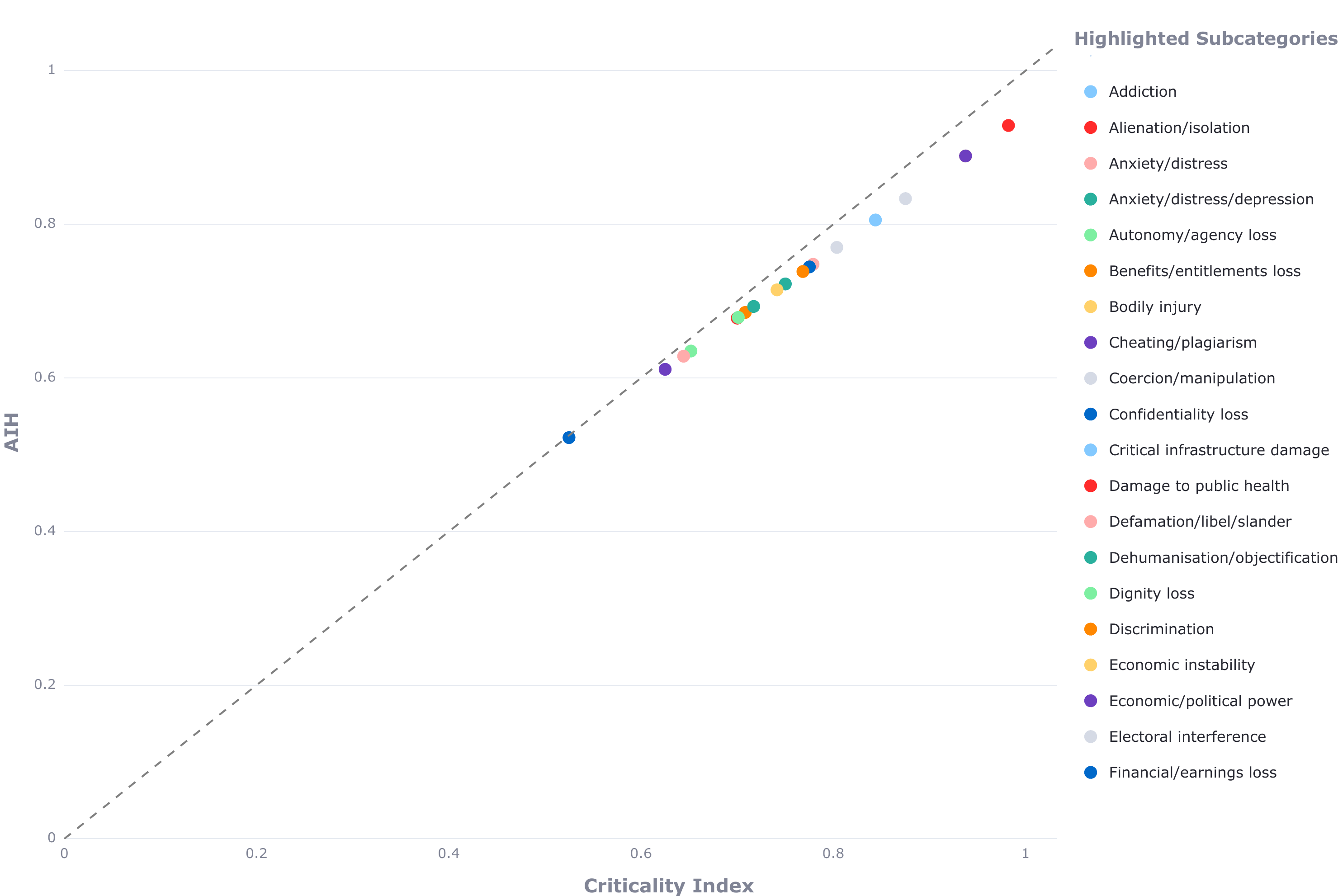}
\caption{Scatter plot comparing mean the $\textit{AIH}$ metric and the Criticality Index for a sample of 20 harm subcategories.}
\label{scattersubs}
\end{figure*}

Figure \ref{scattersubs} compares the $\textit{AIH}$ metric and the Criticality Index for the ordering we used in our initial experiment. A linear correlation is observed, indicating that both methods identify similar patterns of harm concentration at the subcategory level while the uniform trend is highilighted even more clearly here. Each point corresponds to a different subcategory.

\newpage

\section{Appendix: Additional experimental results for sensitivity analysis}\label{lsec:appendix-plots}

\begin{figure*}
\centering
\includegraphics[width=0.7\textwidth]{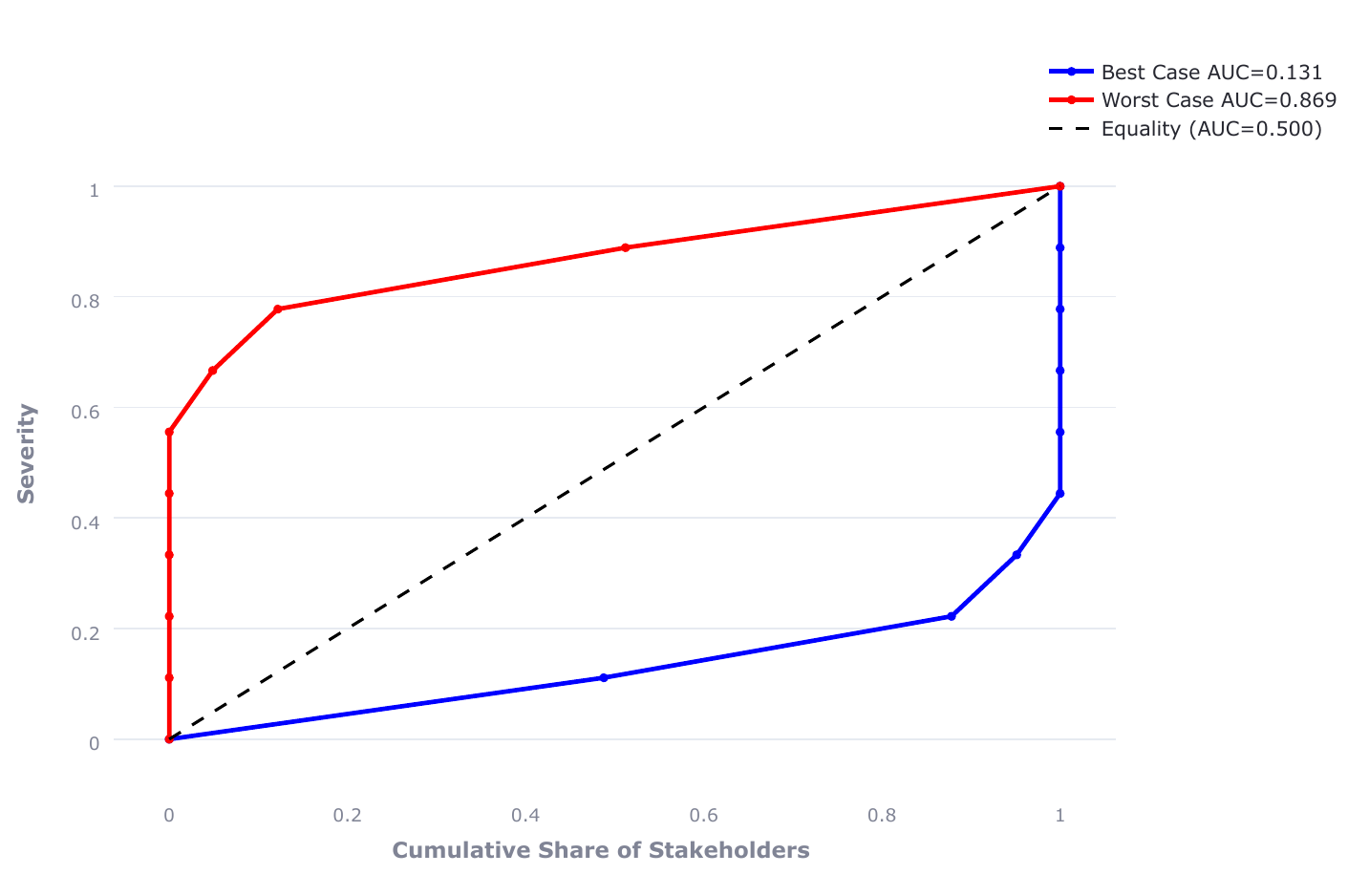}
\caption{$\textit{AIH}$ boundaries as derivative Lorenz Curves for the category "Political \& economic", which appears to have the highest harm concentration among the experiments.}
\label{gba}
\end{figure*}

\begin{figure*}
\centering
\includegraphics[width=0.6\textwidth]{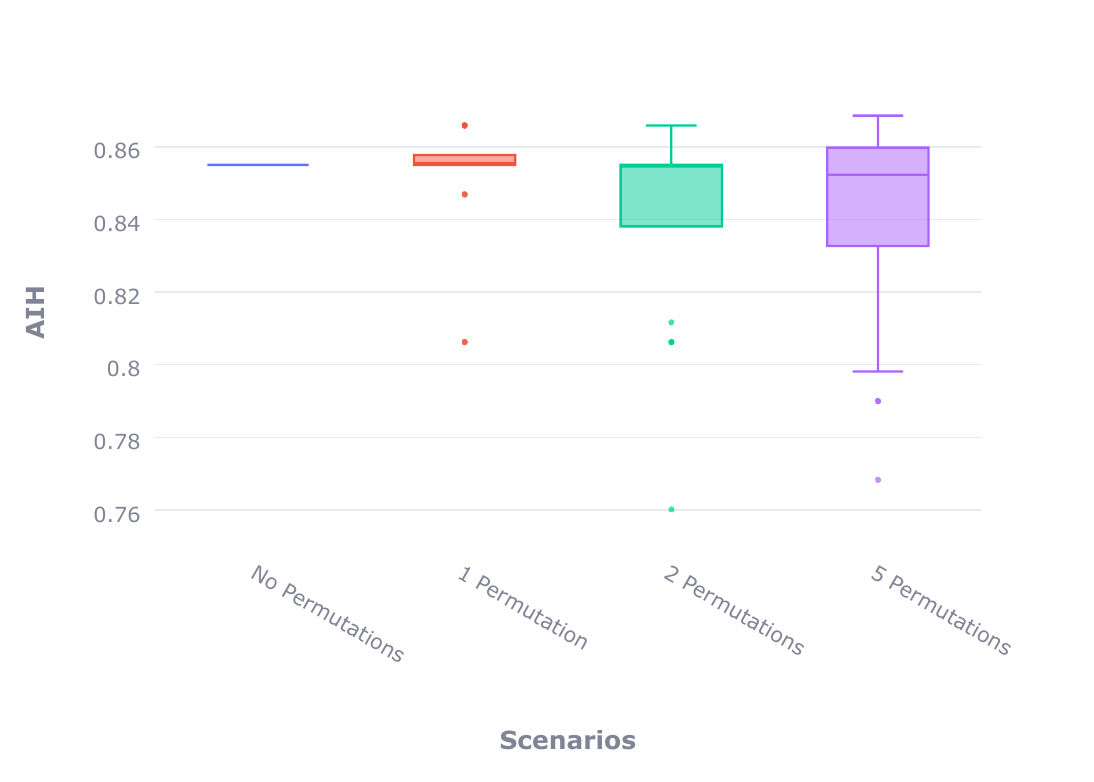}
\caption{Boxplots of the $\textit{AIH}$ for a selected Political \& Economic subcategory, across scenarios.}
\label{fig:drill_boxplot_pe}
\end{figure*}

\begin{figure*}
\centering
\includegraphics[width=0.8\textwidth]{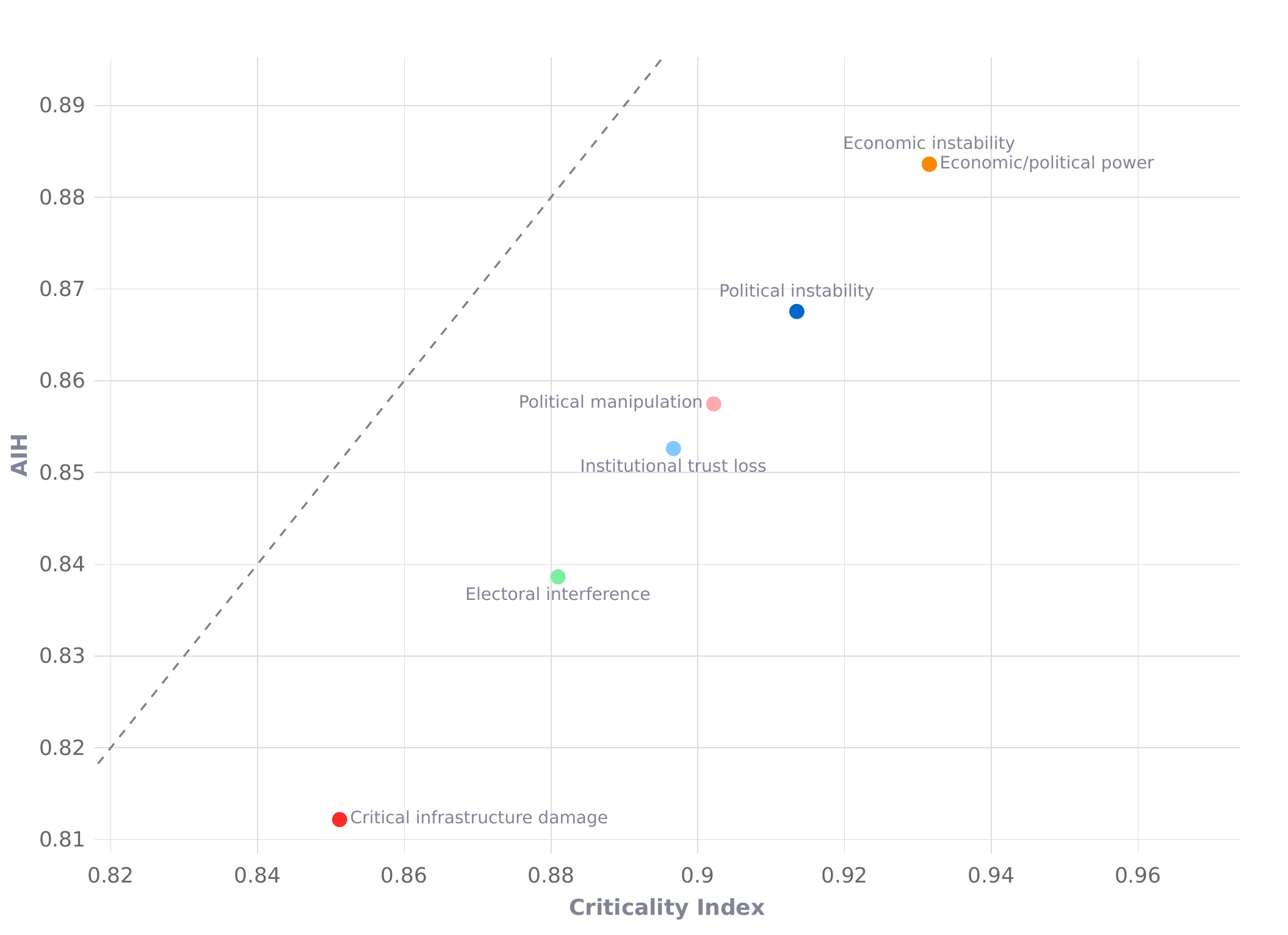}
\caption{Scatter of mean $\textit{AIH}$ vs. Criticality Index for each Political \& Economic subcategory (original ordering).  The dashed 45° line denotes perfect agreement between the two measures.}
\label{fig:drill_scatter_pe}
\end{figure*}

Figure \ref{gba} presents the best-case and worst-case derivative Lorenz curves for the \textit{Political \& economic} harm category. The wide gap between the two curves suggests high sensitivity in this category’s harm inequality when severity assignments vary. This indicates that harm distribution within this category is highly dependent on how severity is assigned to stakeholders, making it a critical area for targeted AI risk assessment. The substantial disparity highlights the potential for severe concentration of harm in such scenarios, reinforcing the need for robust mitigation strategies in AI governance and policy-making. 

To illustrate that robustness holds even within the most unequal category, we present “Political \& Economic”. Figure~\ref{fig:drill_boxplot_pe} shows the $\textit{AIH}$ distribution under No Permutations and under 1, 2, and 5 random swaps.  Figure~\ref{fig:drill_scatter_pe} plots every Political \& Economic subcategory’s mean $\textit{AIH}$ against its CI under the original ordering.

In Figure~\ref{fig:drill_scatter_pe}, the subcategory \emph{Economic/political power} and \emph{Economic instability} indicate the highest concentrations of harm when changing the order of two random severities. At the opposite end, \emph{Critical infrastructure damage} has the lowest $\textit{AIH}$ and CI. The remaining subcategories, \emph{Electoral interference}, \emph{Institutional trust loss}, \emph{Political manipulation} and \emph{Political instability}, all lie between these extremes and fall close to the 45 degree reference line, confirming that both indices produce consistent subcategory rankings within this top-ranked category.

\begin{figure*}
\centering
\includegraphics[width=0.7\textwidth]{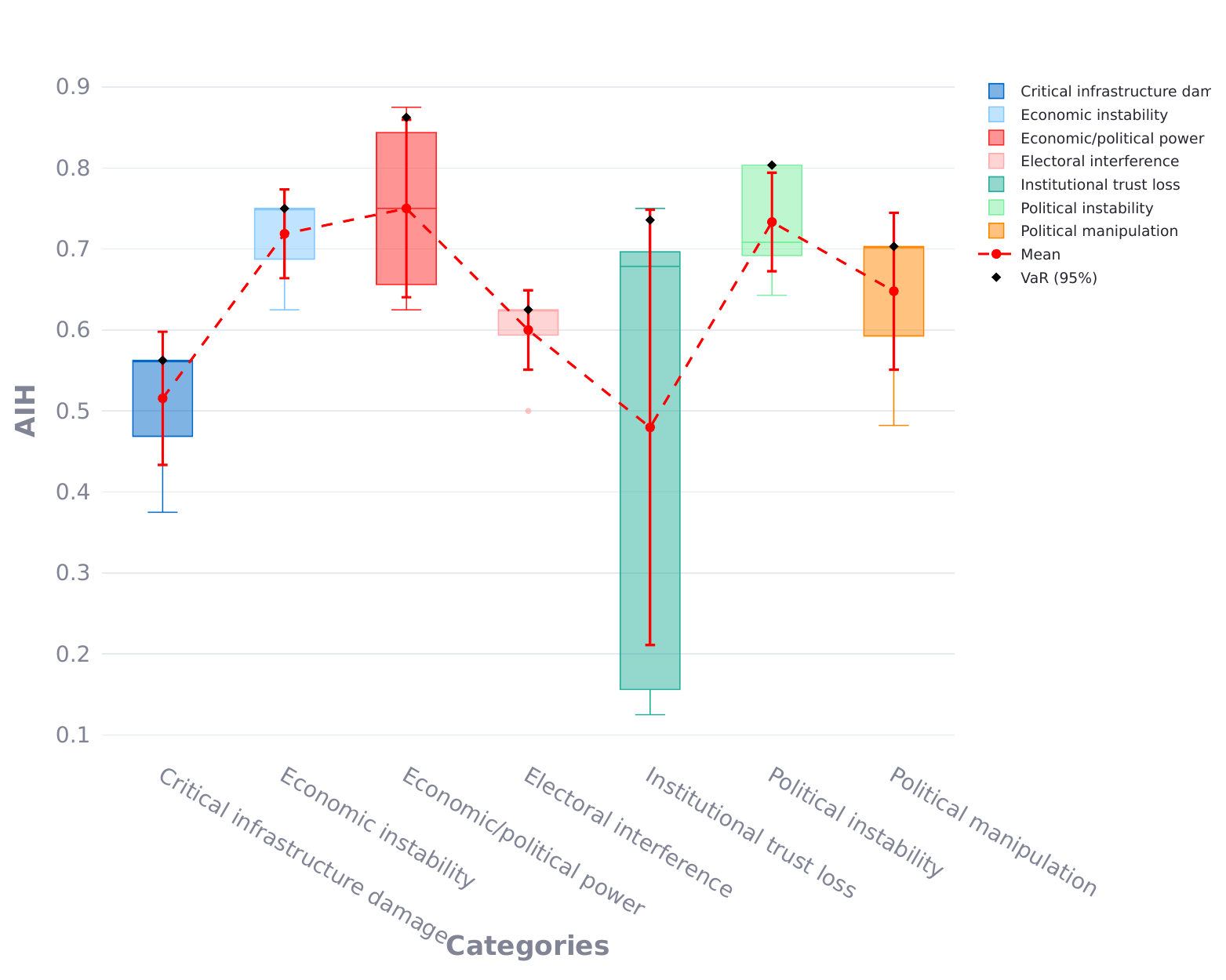}
\caption{$\textit{AIH}$ Boxplots for each Political \& Economic subcategory.}
\label{fig:boxplotspe}
\end{figure*}

Figure~\ref{fig:boxplotspe} presents the distribution of the $\textit{AIH}$ metric across all sensitivity scenarios for each of the seven “Political \& Economic” subcategories.  The \emph{Economic/political power} subcategory clearly stands out with the highest median value, around 0.77 indicating both strongly concentrated harms, while \emph{Economic} and \emph{Political instability} also show relatively high medians. \emph{Institutional trust loss} reveals the largest interquartile range, revealing greater variability under different severity permutations. In contrast, \emph{Critical infrastructure damage} has the lowest median. \emph{Electoral interference} exhibits the smallest interquartile box, which means its $\textit{AIH}$ values remain consistent across all sensitivity scenarios; this stability indicates that shifts in the relative severity ranking have minimal impact on how concentrated these harms appear among stakeholders. 

The findings collectively confirm both the discriminative power and robustness of the \(\mathit{AIH}\) metric across all experiments.  In the “Political \& Economic” category, the large spread between extreme Lorenz curves underscores its potential for very high harm concentration under adversarial severity assignments.  Drilling down to subcategories, \emph{Economic/political power} emerges with the highest median \(\mathit{AIH}\), indicating severe and sensitive concentration of harms, whereas \emph{Critical infrastructure damage} consistently falls at the bottom, reflecting a more uniform impact across stakeholders.  Importantly, \emph{Electoral interference} displays a narrow boxplot, illustrating that its harm distribution is predictably concentrated and largely insensitive to minor perturbations in severity ranking.  Across all subcategories and sensitivity scenarios, the relative ordering of harm importance remains stable (Spearman’s ($\rho \ge 0.97$)), and the category rankings only exhibit modest shifts even when up to 80\% of annotations are randomly removed.  These findings validate that \(\mathit{AIH}\) reliably highlights which harms demand priority attention, combining high concentration, stability under ranking uncertainty, and resilience to incomplete data, thereby fulfilling the goals of a stakeholder‐centric, ordinal harm assessment framework.

\end{document}